# Machine learning-based clinical prediction modeling

*A practical guide for clinicians*

Julius M. Kernbach, MD

*Neurosurgical Artificial Intelligence Lab Aachen (NAILA), RWTH University Aachen, Germany*

&

Victor E. Staartjes, BMed

*Machine Intelligence in Clinical Neuroscience (MICN) Laboratory, University Hospital Zurich, Switzerland*

Contents:

# Machine learning-based clinical prediction modeling
## Part I: Introduction and general principles

*Julius M. Kernbach[1], *MD*; Victor E. Staartjes[2], *BMed*

[1] Neurosurgical Artificial Intelligence Laboratory Aachen (NAILA), Department of Neurosurgery, RWTH Aachen University Hospital, Aachen, Germany
[2] Machine Intelligence in Clinical Neuroscience (MICN) Laboratory, Department of Neurosurgery, University Hospital Zurich, Clinical Neuroscience Center, University of Zurich, Zurich, Switzerland

Corresponding Author
Victor E. Staartjes, BMed
Machine Intelligence in Clinical Neuroscience (MICN) Laboratory
Department of Neurosurgery, University Hospital Zurich
Clinical Neuroscience Center, University of Zurich
Frauenklinikstrasse 10
8091 Zurich, Switzerland
Tel +41 44 255 2660
Fax +41 44 255 4505
Web www.micnlab.com
E-Mail victoregon.staartjes@usz.ch

* J.M. Kernbach and V.E. Staartjes have contributed equally to this series, and share first authorship.

Abstract

As analytical machine learning tools become readily available for clinicians to use, the understanding of key concepts and the awareness of analytical pitfalls are increasingly required for clinicians, investigators, reviewers and editors, who even as experts in their clinical field, sometimes find themselves insufficiently equipped to evaluate machine learning methodologies. In this section, we provide explanations on the general principles of machine learning, as well as analytical steps required for successful machine learning-based predictive modelling, which is the focus of this series. In particular, we define the terms machine learning, artificial intelligence, as well as supervised and unsupervised learning, continuing by introducing optimization, thus, the minimization of an objective error function as the central dogma of machine learning. In addition, we discuss why it is important to separate predictive and explanatory modelling, and most importantly state that a prediction model should not be used to make inferences. Lastly, we broadly describe a classical workflow for training a machine learning model, starting with data pre-processing and feature engineering and selection, continuing on with a training structure consisting of a resampling method, hyperparameter tuning, and model selection, and ending with evaluation of model discrimination and calibration as well as robust internal or external validation of the fully developed model. Methodological rigor and clarity as well as understanding of the underlying reasoning of the internal workings of a machine learning approach are required, otherwise predictive applications despite being strong analytical tools are not well accepted into the clinical routine.

## Introduction

For millennia, the fascination with predicting the future has intrigued humanity. In medicine, numerous risk stratification scores such as the CURB-65 score for mortality risk in community-acquired pneumonia[31], the $CHA_2DS_2$-VASc Score for stroke risk in atrial fibrillation[32], or the modified Fisher scale for vasospasm risk after subarachnoid hemorrhage[15] are used on a daily basis to forecast future events. These risk classification schemes allow stratification of patients into different, broad risk groups - although they are often quite limited in their predictive performance and do not allow personalized predictions for an individual patient. In trying to predict the future, medical professionals are essentially attempting the impossible, as the vast majority of outcomes of interest in clinical medicine are governed by countless minute variables, so that any prediction will always remain only a model of reality, and can take into account only so many factors. This is also evident in the sometimes poor accuracy of experienced clinicians' outcome predictions, often massively underestimating the likelihood of adverse events [36, 39, 40].

Advances in both statistical modeling techniques as well as in computing power over the last few decades have enabled the rapid rise of the field of data science, and improved patient counseling [35, 41]. Even in the field of neurosurgery, ML has been increasingly applied over the years, as evidenced by the sharp rise in publications on machine learning and neurosurgery indexed in PubMed / MEDLINE since the 2000s (Figure 1). While the history of ML applications to the field of neurosurgery is rather compressed into the past two decades, some early efforts have been made as early as the late 1980s. Disregarding other uses of AI and ML – such as advanced histopathological or radiological diagnostics – and focusing on predictive analytics, in 1989 Mathew et al.[33] published a report in which they applied a fuzzy logic classifier to 150 patients, and were able to predict whether disc prolapse or bony nerve entrapment were present based on clinical findings. In 1998, Grigsby et al.[19] were able to predict seizure freedom after anterior temporal lobectomy using neural networks based on EEG and MRI features, using data from 87 patients. Similarly, in 1999, Arle et al. [4] applied neural networks to 80 patients to predict seizures after epilepsy surgery. Soon, and especially since 2010, a multitude of publications followed, applying ML to clinical outcome prediction in all subspecialties of the neurosurgical field [6, 43, 44]. While clinical prediction modeling has certainly been the by far most common application of machine learning in neurosurgery, other exciting

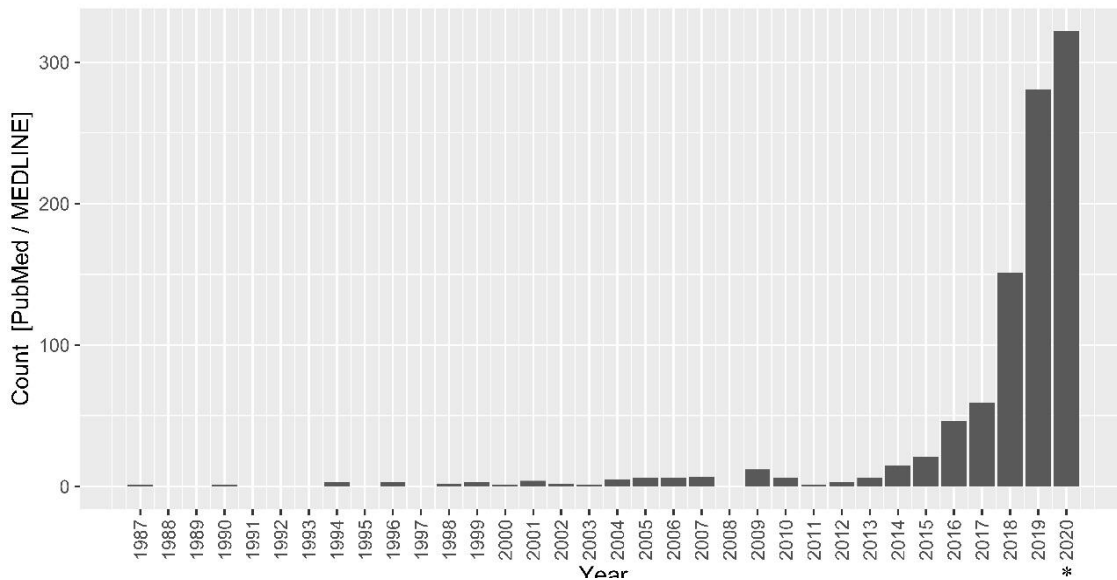

Figure 1 Development of publication counts on machine learning in neurosurgery over the years. Counts were arrived at by searching "neurosurgery" AND ("machine learning" OR "artificial intelligence") on PubMed / MEDLINE. * Projected data

including artificial intelligence (AI) and machine learning (ML) [27]. Along with the broader application of epidemiological principles and larger sample sizes ("big data"), this has led to broad adoption of statistical prediction modelling in clinical practice and research. Clinical prediction models integrate a range of input variables to predict a specific outcome in the future and can aid in evidence-based decision-making applications in neurosurgery such as e.g. in image recognition [49, 52], natural language processing [42], radiomic feature extraction [12, 25], EEG classification [53], and continuous data monitoring [41] should not be disregarded.

Today, ML and other statistical learning techniques have become so easily accessible to anyone with a computer and internet access, that it has become of paramount importance to ensure correct methodology. Moreover, there has been a major "hype" around the terms ML and AI in recent years. Because of their present-day low threshold to accessibility, these techniques can easily be misused and misinterpreted, without intent to do so. For example, it is still common to see highly complex and data-hungry algorithms such as deep neural networks applied to very small datasets, to see overtly overfitted or imbalanced models, or to see models trained for prediction that are then used to "identify risk factors" (explanation or inference). Especially in clinical practice and in the medicolegal arena, it is vital that clinical prediction models intended to be implemented into clinical practice are developed with methodological rigor, and that they are well-validated and generalizable. This series aims to bring the general principles of ML for clinical outcome prediction closer to clinical neuroscientists, and to provide a practical approach to classification and regression problems that the reader may apply to their own research. At this point, it is important to stress that the concepts and methods presented herein are intended as an entry-level guide to ML for clinical outcome prediction, presenting one of many valid approaches to clinical prediction modeling, and thus does not encompass all the details and intricacies of the field. Further reading is recommended, including but not limited to Max Kuhn's "Applied Predictive Modeling" [28] and Ewout W. Steyerberg's "Clinical Prediction Models" [48].

This first part focuses on defining the terms ML and AI in the context of predictive analytics, and clearly describing their applications in clinical medicine. In addition, some of the basic concepts of machine intelligence are discussed and explained. Part II goes into detail about common problems when developing clinical prediction models: What overfitting is and how to avoid it to arrive at generalizable models, how to select which input features are to be included in the final model (feature selection) or how to simplify highly dimensional data (feature reduction). We also discuss how data splits and resampling methods like cross-validation and the bootstrap can be applied to validate models before clinical use. Part III touches on several topics including how to prepare your data correctly (standardization, one-hot encoding) and evaluate models in terms of discrimination and calibration, and points out some recalibration methods. Some other points of significance and caveats that the reader may encounter while developing a clinical prediction model are discussed: sample size, class imbalance, missing data and how to impute it, extrapolation, as well as how to choose a cutoff for binary classification. Parts IV and V present a practical approach to classification and regression problems, respectively. They contain detailed instructions along with a downloadable code for the R statistical programming language, as well as a simulated database of Glioblastoma patients that allows the reader to code in parallel to the explanations. This section is intended as a scaffold upon which readers can build their own clinical prediction models, and that can easily be modified. Furthermore, we will not in detail explain the workings of specific ML algorithms such as generalized linear models, support vector machines, neural networks, or stochastic gradient boosting. While it is certainly important to have a basic understanding of the specific algorithms one applies, these details can be looked up online [37] and detailed explanations of these algorithms would go beyond the scope of this guide. The goal is instead to convey the basic concepts of ML-based predictive modeling, and how to practically implement these.

## Machine Learning - Definitions

As a field of study, ML in medicine is positioned between statistical learning and advanced computer science, and typically evolves around *learning problems*, which can be conceptually defined as optimizing a performance measure on a given task by learning through training experience on prior data. A ML algorithm inductively learns to automatically extract patterns from data to generate insights [21, 24] without being explicitly programmed. This makes ML an attractive option to predict even complex phenomena without pre-specifying an a-priori theoretical model. ML can be used to leverage the full granularity of the data richness enclosed in the *Big Data* trend. Both the complexity and dimensionality of modern medical data sets are constantly increasing and nowadays comprise many variables per observation, much so that we speak of "wide data" with generally more variables (in ML lingo called *features*) than observations (samples) [22, 51]. This has given rise to the so-called '*omics*' sciences including radiomics and genomics [2, 30, 50]. The sheer complexity and volume of data ranging from hundreds to thousands of variables at times exceeds human comprehension, but combined with increased computational power enables the full potential of ML.[37, 54].

With the exponential demand of AI and ML in modern medicine, a lot of confusion was introduced regarding the separation of these two terms. AI and ML are frequently used interchangeably. We define ML as subset of AI – to quote Tom Mitchell – ML "is the study of computer algorithms that allow computer programs to automatically improve through experience" [34], involving the concept of "learning" discussed earlier. In contrast, AI is philosophically much vaster, and can be defined as an ambition to enable computer programs to behave in a human-like nature. That is, showing a certain human-like intelligence. In ML, we learn and optimize a an algorithm from data for maximum performance on a certain learning task. In AI, we try to emulate natural intelligence, to not only learn but apply the gained knowledge to make elaborate decisions and solve complex problems. In a way, ML can thus be considered a technique towards realizing (narrow) AI. Ethical considerations on the "AI doctor" are far-reaching [26, 46], while the concept of a clinician aided by ML-based tools is well-accepted.

The most widely used ML methods are either supervised or unsupervised learning methods, with the exceptions of semi-supervised methods and reinforcement learning [19, 22]. In supervised learning, a set of input variables are used as training set, e.g. different meaningful variables such as age, gender, tumour grading, or functional neurological status to predict a known target variable ("label"), e.g. overall survival. The ML method can then learn the pattern linking input features to target variable, and based on that enable the prediction of new data points – hence, *generalize* patterns beyond the present data. We can train a ML model for survival prediction based on a retrospective cohort of brain tumour patients, since we know the individual length of survival for each patient of the cohort. Therefore, the target variable is *labelled*, and the machine learning-paradigm *supervised*. Again, the actually chosen methods can vary: Common models include support vector machines (SVMs), as example of a *parametric* approach, or the *k*-nearest neighbour (KNN) algorithm as a *nonparametric* method [11]. On the other hand, in *unsupervised* learning, we generally deal with *unlabelled* data with the assumption of the structural coherence. This can be leveraged in clustering, which is a subset of unsupervised learning encompassing many different methods, e.g. hierarchical clustering or *k*-means clustering [3, 21]. The observed data is partitioned into clusters based on a measure of similarity regarding the structural architecture of the data. Similarly, dimensionality reduction methods – including principal component analysis (PCA) or autoencoders - can be applied to derive a low-dimensional representation explicitly from the present data [21, 35].

A multitude of diverse ML algorithms exist, and sometimes choosing the "right" algorithm for a given application can be quite confusing. Moreover, based on the so-called *no free lunch theorem* [56] no single statistical algorithm or model can generally be considered superior for all circumstances. Nevertheless, ML algorithms can vary greatly based on the (a) representation of the candidate algorithm, (b) the selected performance metric and (c) the applied optimization strategy[14, 21, 24]. Representation refers to the learner's hypothesis space of how they formally deal with the problem at hand. This includes but is not limited to instance-based learners, such as KNN, which instead of performing explicit generalization compares new observations with similar instances observed during training.[5] Other representation spaces include hyperplane-based models, such as logistic regression or naïve Bayes, as well as rule-based learners, decision trees or complex neural networks, all of which are frequently leveraged in various ML problems across the neurosurgical literature[9, 43]. The evaluated performance metrics can vary greatly, too. Performance evaluation and reporting play a pivotal role in predictive analytics (c.f. Part III). Lastly, the applied ML algorithm is *optimized* by a so-called objective function such as greedy search or unconstrained continuous optimization options, including different choices of gradient descent [7, 38]. Gradient descent represents the most common optimization strategy for neural networks and can take different forms, e.g. batch-("vanilla"), stochastic- or mini-batch gradient descent[38]. We delve deeper into optimization to illustrate how it is used in learning.

## Optimization – The Central Dogma of Learning Techniques

At the heart of nearly all ML and statistical modeling techniques used in data science lies the concept of *optimization*. Even though optimization is the backbone of algorithms ranging from linear and logistic regression to neural networks, it is not often stressed in the non-academic data science space. Optimization describes the process of iteratively adjusting parameters to improve performance. Every optimization problem can be decomposed into three basic elements: First, every algorithm has *parameters* (sometimes called *weights*) that govern how the values of the input variables lead to a prediction. In linear and logistic regression for example, these parameters include the coefficients that are multiplied with the input variable values, as well as the intercept. Second, there may be realistic *constraints* within which the parameters, or their combinations, must fall. While simple models such as linear and logistic regression often do not have such constraints, other ML algorithms such as support vector machines or *k*-means clustering do. Lastly and importantly, the optimization process is steered by evaluating a so-called *objective function* that assesses how well the current iteration of the algorithm is performing. Commonly, these objective functions are *error* (also called *loss*) functions, describing the deviation of the predicted values from the true values that are to be predicted. Thus, these error functions must be *minimized*. Sometimes, you may choose to use indicators of performance, such as accuracy, which conversely need to be *maximized* throughout the optimization process.

| Classical/Inferential Statistics | Statistical/Machine Learning |
|---|---|
| **Explanatory modeling** | **Predictive modeling** |
| An a priori chosen theoretical model is applied to data in order to test for causal hypotheses. | The process of applying a statistical model or data mining algorithm to data for the purpose of predicting new or future observations. |
| **Focus on in-sample estimates** | **Focus on out-of-sample estimates** |
| Goal: to confirm the existence of an effect in the entire data sample. Often using significance testing. | Goal: Use the best performing model to make new prediction for single new observations. Often using resampling techniques. |
| **Focus on model interpretability** | **Focus on model performance** |
| The model is chosen apriori, while models with intrinsic means of interpretability are preferred, e.g. a GLM, often parametric with a few fixed parameters. | Different models are applied and the best performing one is selected. Models tend to be more flexible and expressive, often non-parametric with many parameters adapting to the present data. |
| Experimental data | Empirical data |
| Long data (n samples > p variables) | Wide data (n samples << p variables) |
| Independent variables | Features |
| Dependent variable | Target variable |
| Learn deductively by model testing | Learn a model from data inductively |

Table 1 A comparison of central concepts in classical / inferential statistics versus in statistical / machine learning.

The optimization process starts by randomly *initializing* all model parameters – that is, assigning some initial value for each parameter. Then, predictions are made on the training data, and the error is calculated. Subsequently, the parameters are adjusted in a *certain* direction, and the error function is evaluated again. If the error increases, it is likely that the direction of adjustment of the parameters was awry and thus led to a higher error on the training data. In that case, the parameter values are adjusted in different directions, and the error function is evaluated again. Should the error decrease, the parameter values will be further modified in these specific directions, until a *minimum* of the error function is reached. The goal of the optimization process is to reach the *global minimum* of the error function, that is, the lowest error that can be achieved through the combination of parameter values within their constraints. However, the optimization algorithm must avoid getting stuck at *local minima* of the error function (see Figure 2).

The way in which the parameters are adjusted after each iteration is governed by an *optimization algorithm*, and approaches can differ greatly. For example, linear regression usually uses the ordinary least square (OLS) optimization method. In OLS, the parameters are estimated by solving an equation for the minimum of the sum of the square errors. On the other hand, *stochastic gradient descent* – which is a common optimization method for many ML algorithms – iteratively adjusts parameters as described above and as illustrated in Figure 2. In stochastic gradient descent, the amount by which the parameters are changed after each iteration (also called *epoch*) is controlled by the calculated derivative (i.e. the slope or *gradient*) for each parameter with respect to the error function, and the *learning rate*. In many models, the learning rate is an important hyperparameter to set, as it controls how much parameters change in each iteration.

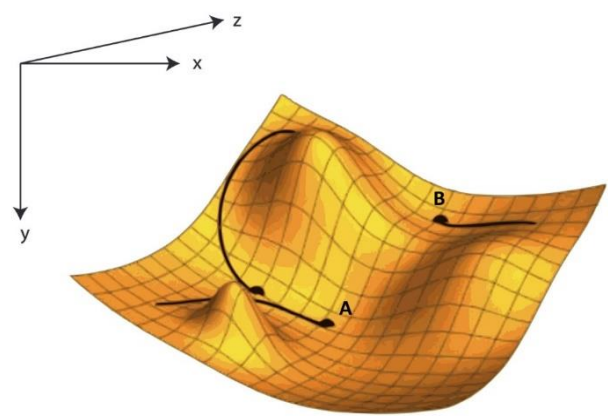

Figure 2 Illustration of an optimization problem. In the x and z dimension, two parameters can take different values. In the y dimension, the error is displayed for different values of these two parameters. The goal of the optimization algorithm is to reach the *global minimum* (A) of the error through adjusting the parameter values, without getting stuck at a *local minimum* (B). In this example, three models are initialized with different parameter values. Two of the models converge at the global minimum (A), while one model gets stuck at a local minimum (B). Illustration by Jacopo Bertolotti. This illustration has been made available under the Creative Commons CC0 1.0 Universal Public Domain Dedication.

On the one hand, small learning rates can take many iterations to converge and make getting stuck at a local minimum more likely – on the other hand, a large learning rate can overshoot the global minimum. As a detailed discussion of the mathematical nature behind different algorithms remains beyond the scope of this introductory series, we refer to popular standard literature such as "Elements of Statistical Learning" by Hastie and Tibshirani [21], "Deep Learning" by Goodfellow et al. [17], and "Optimization for Machine Learning" by Sra et al.[47]

## Explanatory Modeling versus Predictive Modeling

The 'booming' of applied ML has generated a methodological shift from *classical statistics* (experimental setting, hypothesis testing, group comparison, inference) to data-driven *statistical learning* (empirical setting, algorithmic modeling comprising ML, AI, pattern recognition).[18] Unfortunately, the two statistical cultures have developed separately over the past decades [8] leading to incongruent evolved terminology and misunderstandings in the absence of an agreed-upon technical theorem (Table 1). This already becomes evident in the basic terminology describing model inputs and outputs: *predictors* or *independent variables* refer to model inputs in classical statistics, while *features* are the commonly used term in ML; outputs, known as *dependent variable* or *response*, are often labeled *target variable* or *label* in ML instead [10]. The duality of language has led to misconceptions regarding the fundamental difference between inference and prediction, as the term *prediction* has frequently been used incompatibly as in-sample correlation instead of out-of-sample generalization [16, 45]. The variation of one variable with a subsequent correlated variable later in time, such as the outcome, in the same group (in-sample correlation) does not imply prediction, and failure to account for this distinction can lead to false clinical decision making[55, 57]. Strong associations between variables and outcome in a clinical study remain averaged estimates of the evaluated patient cohort, which does not necessarily enable predictions in unseen new patients. To shield clinicians from making wrong interpretations, we clarify the difference between explanatory modeling and predictive modeling, and highlight the potential of ML for strong predictive models.

Knowledge generation in clinical research has nearly exclusively been dominated by classical statistics with the focus on *explanatory modeling (EM)*[45]. In carefully designed experiments or clinical studies, a constructed theoretical model, e.g. a regression model, is applied to data in order to test for causal hypotheses. Based on theory, a model is chosen a priori, combining a fixed number of experimental variables, which are under the control of the investigator. Explicit model assumptions such as the gaussian distribution assumption are made, and the model, which is believed to represent the true *data generating process*, is evaluated for the entire present data sample based on hypothesis and significance testing ("inference"). In such association-based modeling, a set of independent variables (X) are assumed to behave according to a certain mechanism ("theory") and ultimately cause an effect measured by the dependent variable (Y). Indeed, the role of *theory* in explanatory modeling is strong and is always reflected in the applied model, with the aim to obtain the most accurate representation of the underlying theory (technically speaking, classical statistics seeks to minimize *bias*). Whether *theory* holds true and the effect actually exists is then confirmed in the data, hence the overall analytical goal is *inference*.

Machine learning-based *predictive modeling* (PM) is defined as the process of applying a statistical model or data mining algorithm to data for the purpose of predicting future observations. In a heuristic approach, ML or PM is applied to *empirical data* as opposed to experimentally controlled data.

As the name implies, the primary focus lays on optimizing the prediction of a target variable (Y) on new observations given their set of features (X). As opposed to explanatory modeling, PM is *forward looking*[45] with the intention of predicting new observations, and hence *generalization beyond the present data* is the fundamental goal of the analysis. In contrast to EM, PM seeks to minimize both *variance* and *bias*[13, 23], occasionally sacrificing the theoretical interpretability for enhanced predictive power. Any underlying method can constitute a predictive model ranging from parametric and rigid models to highly flexible non-parametric and complex models. With a minimum of a-priori specifications, a model is then heuristically derived from the data[1, 29]. The true data generating process lays in the data, and is inductively learned and approximated by ML models.

## Workflow for Predictive Modeling

In clinical predictive analytics, *generalization* is our ultimate goal. To answer different research objectives, we develop, test and evaluate different models for the purpose of clinical application (for an overview see https://topepo.github.io/caret/available-models.html). Many research objectives in PM can be framed either as the prediction of a continuous endpoint (regression) such as progression-free survival measured in months or alternatively as the prediction of a binary endpoint. (classification), e.g. survival after 12 months as a dichotomized binary. Most continuous variables can easily be reduced and dichotomized into binary variables, but as a result data granularity is lost. Both regression and classification share a common analytical workflow with difference in regard to model evaluation and reporting (c.f. *Part IV Classification problems* and *V Regression problems* for a detailed discussion). An adaptable pipeline for both regression and classification problems is demonstrated in Part IV and V. Both sections contain detailed instructions along with a simulated dataset of 10'000 patients with glioblastoma and the code based on the statistical programming language R, which is available as open-source software.

For a general overview, a four-step approach to PM is proposed (Figure 3): First and most important (1) all data needs to be pre-processed. ML is often thought of as *letting data do the heavy lifting,* which in part is correct, however the raw data is often not suited to learning well in its current form. A lot of work needs to be allocated to preparing the input data including data cleaning and pre-processing (imputation, scaling, normalization, encoding) as well as *feature engineering* and *selection*. This is followed by using (2) resampling techniques such as *k*-fold cross validation (c.f. *Part II: generalization and overfitting)* to train different models and perform hyperparameter tuning. In a third step (3), the different models are compared and evaluated for generalizability based on a chosen out-of-sample performance measure in an independent testing set. The best performing model is ultimately selected, the model's out-of-sample calibration assessed (c.f. *Part III: Evaluation and points of significance*), and, in a fourth step (4) the model is externally validated – or at least prospectively internally validated – to ensure clinical usage is safe and generalizable across locations, different populations and end users (c.f. *Part II Generalization and overfitting*). The European Union (EU) and the Food and Drug Administration (FDA) have both set standards for classifying machine learning and other software for use in healthcare, upon which the extensiveness of validation that is required before approved introduction into clinical practice is based. For example, to receive the CE mark for a clinical decision support (CDS) algorithm – depending on classification – the EU requires compliance with ISO 13485 standards, as well as a clinical evaluation report (CER) that includes a literature review and clinical testing (validation).[20]

Conclusion
We appear to be at the beginning of an accelerated trend towards data-driven decision-making in biomedicine enabled by a transformative technology – machine learning [24]. Given the ever-growing and highly complex "big data" biomedical datasets and increases in computational power, machine learning approaches prove to be highly successful analytical strategies towards a patient-tailored approach regarding diagnosis, treatment choice and outcome prediction. Going forward, we expect that training neuroscientists and clinicians in the concepts of machine learning will undoubtedly be a corner stone for the advancement of individualized medicine in the realm of precision medicine. With the series "*Machine learning-based clinical prediction modeling*", we aim to provide both a conceptual and practical guideline for predictive analytics in the clinical routine to strengthen every clinician's competence in modern machine learning techniques.

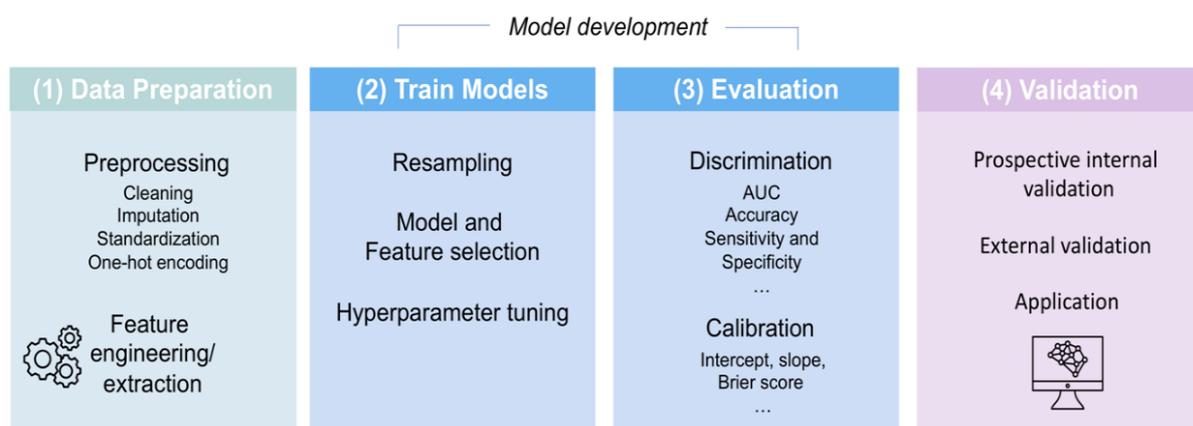

Figure 3 A four-step predictive modeling workflow. (1) Data preparation includes cleaning and featurization of the given raw data. Data pre-processing combines cleaning and outlier detection, missing data imputation, the use of standardization methods, and correct feature encoding. The pre-processed data is further formed into features – manually in a process called *feature engineering* or automatically deduced by a process called *feature extraction*. In the training process (2) resampling techniques such as *k*-fold cross validation are used to train and tune different models. Most predictive features are identified in a *feature selection* process. (3) Models are compared and evaluated for generalizability in an independent testing set. The best performing model is selected, and out-of-sample discrimination and calibration are assessed. (4) The generalizing model is prospectively internally and externally validated to ensure safe clinical usage across locations and users.

Disclosures

Funding: No funding was received for this research.

Conflict of interest: All authors certify that they have no affiliations with or involvement in any organization or entity with any financial interest (such as honoraria; educational grants; participation in speakers' bureaus; membership, employment, consultancies, stock ownership, or other equity interest; and expert testimony or patent-licensing arrangements), or non-financial interest (such as personal or professional relationships, affiliations, knowledge or beliefs) in the subject matter or materials discussed in this manuscript.

Ethical approval: All procedures performed in studies involving human participants were in accordance with the ethical standards of the 1964 Helsinki declaration and its later amendments or comparable ethical standards.

Informed consent: No human or animal participants were included in this study.

# Machine learning-based clinical prediction modeling
## Part II: Generalization and Overfitting

*Julius M. Kernbach[1], *MD*; Victor E. Staartjes[2], *BMed*

[1] Neurosurgical Artificial Intelligence Laboratory Aachen (NAILA), Department of Neurosurgery,
RWTH Aachen University Hospital, Aachen, Germany
[2] Machine Intelligence in Clinical Neuroscience (MICN) Laboratory, Department of Neurosurgery,
University Hospital Zurich, Clinical Neuroscience Center, University of Zurich, Zurich, Switzerland

Corresponding Author
Victor E. Staartjes, BMed
Machine Intelligence in Clinical Neuroscience (MICN) Laboratory
Department of Neurosurgery, University Hospital Zurich
Clinical Neuroscience Center, University of Zurich
Frauenklinikstrasse 10
8091 Zurich, Switzerland
Tel +41 44 255 2660
Fax +41 44 255 4505
Web www.micnlab.com
E-Mail victoregon.staartjes@usz.ch

* J.M. Kernbach and V.E. Staartjes have contributed equally to this series, and share first authorship.

Abstract

In this section, we review the concept of *overfitting*, which is a well-known concern within the machine learning community, but less established in the clinical community. Overfitted models may lead to inadequate conclusions that may wrongly or even harmfully shape clinical decision- making. Overfitting can be defined as the difference among discriminatory training and testing performance - while it is normal that out-of-sample performance is equal to or ever so slightly worse than training performance for any adequately fitted model, a massive worse out-of-sample performance suggests relevant overfitting. As strategies to combat overfitting, we delve into resampling methods, and specifically recommend the use of *k*-fold cross validation and the bootstrap to arrive at realistic estimates of out-of-sample error during training. Also, we encourage the use of regularization techniques such as L1 or L2 regularization, and to choose an appropriate level of algorithm complexity for the type of dataset used. To further prevent overfitting, the concept of data leakage or data contamination is addressed - when information about the test data leaks into the training data. Also, the importance of external validation to assess true out-of-sample performance and to - upon successful external validation - release the model into clinical practice is discussed. Finally, for highly dimensional datasets, the concepts of feature reduction using principal component analysis (PCA) as well as feature elimination using recursive feature elimination (RFE) are elucidated.

## Introduction

In the first part of this review series, we have discussed general and important concepts of machine learning (ML) and presented a four-step workflow for machine learning-based predictive pipelines. However, many regularly faced challenges, which are well-known within the ML community, are less established in the clinical community. One common source of trouble is *overfitting*. It is a common pitfall in predictive modelling, whereby the model not only fits the true underlying relationship of the data but also fits the individual biological or procedural noise associated with each observation. Dealing with overfitting remains challenging in both regression and classification problems. Erroneous pipelines or ill-suited applied models may lead to drastically inflated model performance, and ultimately cause unreliable and potentially harmful clinical conclusions. We discuss and illustrate different strategies to address overfitting in our analyses including *resampling methods*, regularization and penalization of model complexity [7]. In addition, we discuss *feature selection* and *feature reduction*. In this section, we review overfitting as potential danger in predictive analytic strategies with the goal of providing useful recommendations for clinicians to avoid flawed methodologies and conclusions (Table 1).

## Overfitting

Overfitting occurs when a given model adjusts too closely to the training data, and subsequently demonstrates poor performance on the testing data (Figure 1). While the model's goodness of fit to the present data sample seems impressive, the model will be unable to make accurate predictions on new observations. This scenario represents a major pitfall in ML. At first, the performance within the training data seems excellent, but when the model's performance is evaluated on the hold-out data ("out-of-sample error") it generalizes poorly. There are various causes of overfitting, some of which are intuitive and easily mitigated. Conceptually, the easiest way to overfit is simply by memorizing observations [2, 17, 33].

| Concept | Explanation |
|---|---|
| Noise | Noise is unexplained and random variation inherent to the data (biological noise) or introduced by variables of no interest (procedural noise, including measurement errors, site variation). |
| Overfitting | Over-learning of random patterns associated with noise or memorization in the training data. Overfitting leads to a drastically decreased ability to generalize to new observations. |
| Bias | Bias quantifies the error term introduced by approximating highly complicated real-life problems by a much simpler statistical model. Models with high bias tend to underfit. |
| Variance | Variance refers to learning random structure irresponsible of the underlying true signal. Models with high variance tend to overfit. |
| Data Leakage / Contamination | Or the concept of "looking at data twice". Overfitting is introduced when observations used for testing also re-occur in the training process. The model then "remembers" instead of learning the underlying association. |
| Model Selection | Iterative process using resampling such as k-fold cross-validation to fit different models in the training set. |
| Model Assessment | Evaluation of a model's out-of-sample performance. This should be conducted on a test set of data that was set aside and not used in training or model selection. The use of multiple measures of performance (AUC, F1 etc.) are recommended. |
| Resampling | Resampling methods fit a model multiple times on different subsets of the training data. Popular methods are k-fold cross-validation and the bootstrap. |
| *K*-Fold Cross Validation | Data is divided in k equally sized folds/sets. Iteratively, k-1 data is used for training and evaluated on the remaining unseen fold. Each fold is used for testing once. |
| LOOCV | LOOCV (leave-one-out cross validation) is a variation of cross-validation. Each observation is left out once, the model is trained on the remaining data, and then evaluated on the held-out observation. |
| Bootstrap | The bootstrap allows to estimate the uncertainty associated with any given model. Typically, in 1000-10000 iterations bootstrapped samples are repetitively drawn with replacement from the original data, the predictive model is iteratively fit and evaluated. |
| Hyperparameter Tuning | Hyperparameters define how a statistical model learns and need to be specified before training. They are model specific and might include regularization parameters penalizing model's complexity (ridge, lasso), number of trees and their depth (random forest), and many more. Hyperparameters can be tuned, that is, iteratively improved to find the model that performs best given the complexity of the available data. |

Table 1 Concept summaries.

We simply remember all data patterns, important patterns as well as unimportant ones. For our training data, we will get an exceptional model fit, and minimal training error by recalling the known observations from memory – implying the illusion of success. However, once we test the model's performance on independent test data, we will observe predictive performance that is no better than random. By over-training on the present data, we end up with a too close fit to the training observations. This fit only partially reflects the underlying true data-generating process, but also includes random noise specific to the training data. This can either be sample-specific noise, both procedural as well as biological, but also the hallucination of unimportant patterns [9]. Appling the overfitted model to new observations will out itself as a out-of-sample performance that is massively worse than the training performance. In this way, the amount of overfitting can be defined as the difference among discriminatory training and testing performance – while it is normal that out-of-sample performance is equal to or ever so slightly worse than training performance for any adequately fitted model, a massive difference suggests relevant overfitting. This is one reason why in-sample model performance should never be reported as evidence for predictive performance. Instead model training and selection should always be performed on a separate train set, and only in the final step should the final model be evaluated on an independent test set to judge true out-of-sample performance.

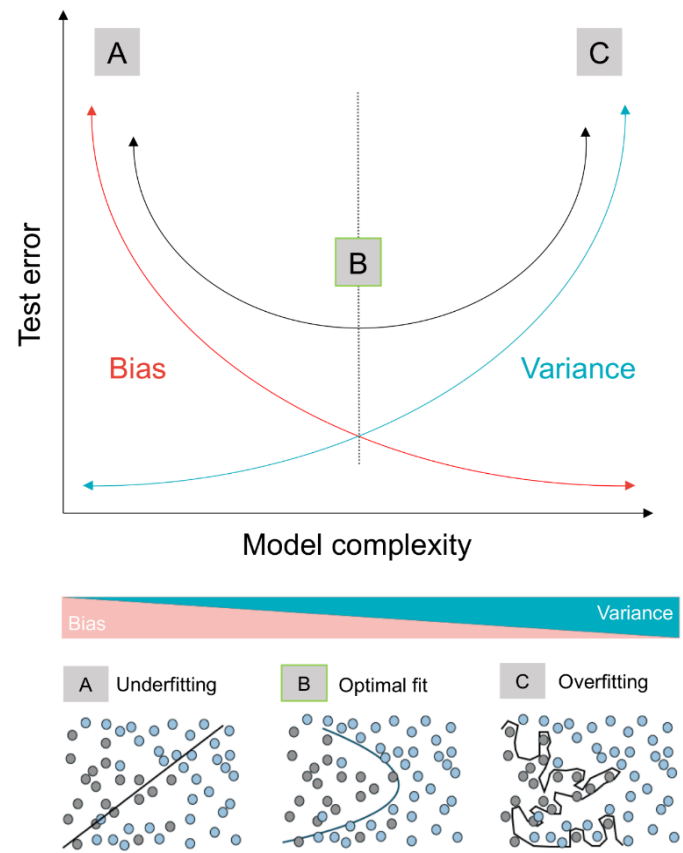

Figure 1 Conceptual visualization of the bias-variance trade-off. A predictive model with *high bias* and *low variance* (A), consistently approximates the underlying data-generating process with a much simpler model (here a hyperplane), and hence result in an underfit solution. (B) A U-shaped decision boundary represents the optimal solution in this scenario, here, both bias and variance are low, resulting in the lowest test error. (C) Applying an overly flexible model results in overfitting. Data quirks and random non-predictive structures that are unrelated to the underlying signal are learned

## The Bias-Variance Trade-Off

In ML we opt to make accurate and generalizable predictions. When the test error is significantly higher than the training error, we can diagnose overfitting. To understand what is going on we can decompose the predictive error into its essential parts *bias* and *variance* [8, 19]. Their competing nature, commonly known under the term *bias-variance trade-off*, is very important and notoriously famous in the machine learning community. Despite its fame and importance, the concept is less prominent within the clinical community. *Bias* quantifies the error term introduced by approximating highly complicated real-life problems by a much simpler statistical model, that is underfitting the complexity of the data-generating process. In other words, a model with high bias tends to consistently learn the wrong response. That by itself does not necessarily need to be a problem, as simple models were often found to perform very well sometimes even better than more sophisticated ones [16]. However, for maximal predictive compacity we need to find the perfect balance between bias and variance. The term *variance* refers to learning random structure irresponsible of the underlying true signal. That is, models with high variance can hallucinate patterns that are not given by the reality of the data. Figure 1 illustrates this in a classification problem. A linear model (Figure 1A, high bias and low variance) applied to class data, in which the frontier between the two classes is not a hyperplane, is unable to induce the underlying true boundary. It will consistently learn the wrong response, that is a hyperplane, despite the more complex true decision boundary and result into 'underfitting' the true data-generating process. On the other extreme, an excessively flexible model with high variance and low bias (Figure 1C) will learn random non-predictive structure that is unrelated to the underlying signal. Given minimally different observations, the overly flexible model fit could drastically change in an instance. The latter complex model would adapt well to all training observations but would ultimately fail to generalize and predict new observations in an independent test set. Neither the extremely flexible nor the insufficiently flexible model is capable of generalizing to new observations.

*Combatting Overfitting: Resampling*
We could potentially collect more data for an independent cohort to test our model, but this would be highly time-consuming and expensive. In rich data situations, we can alternatively split our sample into a data set for training and a second set for testing (or hold-out set) to evaluate the model's performance in new data (i.e., the model's out-of-sample performance) more honestly. We would typically use a random 80%/20% split for training and testing (while remaining class balance within the training set, c.f. *Part III*). Because we often lack a sufficiently large cohort of patients to simply evaluate generalization performance using data splits, we need to use a less data-hungry but equally efficient alternatives. The gold standard and popular approach in machine learning to address overfitting is to evaluate the model's generalization ability via *resampling methods*.[29] Some of these resampling methods – particularly the bootstrap – have already long been used in inferential statistical analysis to generate measures of variance.[6] Resampling methods are an indispensable tool in today's modern data science and include various forms of *cross-validation* [12, 17]. All forms have a common ground: they involve splitting the available data iteratively into a non-overlapping train and test set. Our statistical model is then refitted and tested for each subset of the train and test data to obtain an estimate of generalization performance. Most modern resampling methods have been derived from the jackknife – a resampling technique developed by Maurice Quenouille in 1949.[27] The simplest modern variation of cross-validation – also based on the jackknife – is known as leave-one-out cross-validation (LOOCV). In LOOCV, the data ($n$) is iteratively divided into two unequal subsets with the train set of $n$-1 observations and the test set containing the remaining one observation. The model is refitted and evaluated on the excluded held-out observation. The procedure is then repeated $n$ times and the test error is then averaged over all iterations. A more popular alternative to LOOCV and generally considered the gold standard is $k$-fold cross validation (Figure 2). The k-fold approach randomly divides the available data into a $k$ amount of nonoverlapping groups, or folds, of approximately equal size. Empirically, $k$=5 or $k$=10 are preferred and commonly used [14]. Each fold is selected as test set once, and the model is fitted on the remaining $k$-1 folds. The average over all fold-wise performances estimates the generalizability of a given statistical model. Within this procedure, importantly, no observation is selected for both training and testing.

This is essential, because, as discussed earlier, predicting an observation that was already learned during training equals memorization, which in turn leads to overfitted conclusions.

Cross-validation is routinely used in both model selection and model assessment. Yet another extremely powerful and popular resampling strategy is the *bootstrap* [10, 15], which allows for the estimation of the accuracy's uncertainty applicable to nearly any statistical method. Here, we obtain new bootstrapped sets of data by repeatedly sampling observations from the original data set *with replacement*, which means any observation can occur more than once in the bootstrapped data sample. Thus, when applying the bootstrap, we repeatedly randomly select $n$ patients from an $n$-sized training dataset, and model performance is evaluated after every iteration. This process is repeated many times - usually with 25 to 1000 repetitions.

*Considerations on Algorithm Complexity*
To avoid over- or underfitting, an appropriate level of model complexity is required.[12, 30] Modulating complexity can be achieved by adding a regularization term, which can be used with any type of predictive model. In that instance, the regularization term is added to favour less-complex models with less room to overfit. As complexity is intrinsically related to the number and magnitude of parameters, we can add a regularization or penalty term to control the magnitude of the model parameters, or even constrain the number of parameters used. There are many different penalties specific to selected models. In a regression setting, we could add either a *L1 penalty* (LASSO, least absolute shrinkage and selection operator), which selectively removes variables form the model, a *L2 penalty* (Ridge or Tikhonov regularization), which shrinks the magnitude of parameters but never fully removes them from the model, or an *elastic net* (combination of L1 and L2) [14, 23, 34]. For neural networks, *dropout* is a very efficient regularization method.[28] Finding the right balance based on regularization, that is, to define how complex a model can be, is controlled by the model's hyperparameters (L1 or L2 penalty term in regression, and many more). Restraining model complexity by adding a regularization term is an example of a model hyperparameter. Typically, hyperparameters are *tuned*, which means that the optimal level is evaluated during model training. Again, it is important to respect the distinction of train and test data. As a simple guideline, we recommend to automate all necessary pre-processing steps including hyperparameter tuning within the chosen resampling approach to ensure none of the above are performed on the complete data set before cross validation [26]. Otherwise, this would result in circularity and inflate the overall predictive performance [22].

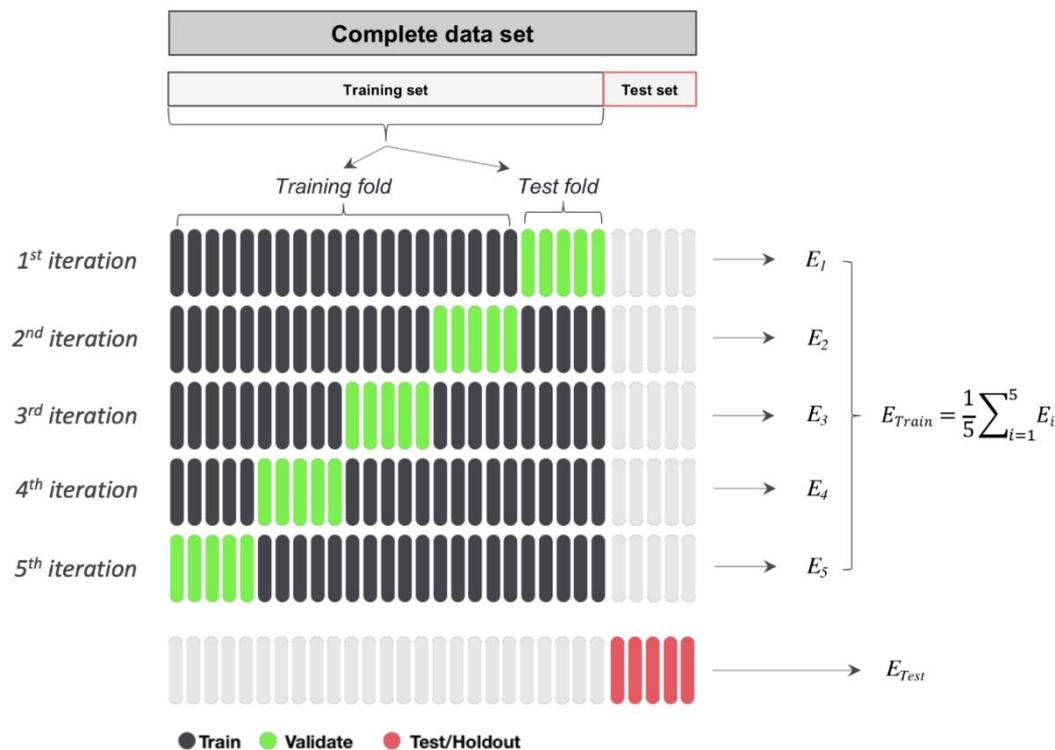

**Figure 2** *K*-fold cross validation with an independent holdout set. The complete dataset is portioned into training data (~80%) and testing data (~20%) before any resampling is applied. Within the training set, *k*-fold cross validation is used to randomly divide the available data into *k*=5 equally sized folds. Iteratively, *k*-1 folds are used to train a chosen model, and the fold-wise performance ($E_i$) is evaluated on the remaining unseen validation fold. These fold-wise performances are averaged, and together, the out-of-sample performance is estimated as E $_{Train}$. When different models are trained, the best performing one is selected and tuned (model selection, hyperparameter tuning) and evaluated on the independent holdout set (or "test set"). The resulting performance E $_{Test}$ is reported and estimates the predictive performance beyond the present data.

*Data Leakage*

Whenever resampling techniques are applied, the investigator has to ensure that *data leakage* or *data contamination* is not accidently introduced. From the standpoint of ML, data contamination – part of the test data leaking into the model-fitting procedure – can have severe consequences, and lead to drastically inflated predictive performance. Therefore, caution needs to be allocated to the clean isolation of train and test data. As a general rule-of-thumb, no feature selection or dimensionality reduction method that involves the outcome measure should be performed on the complete data set before cross validation or splitting. This would open doors for procedural bias, and raise concerns regarding model validity. Additionally, nested cross validation should be used in model selection and hyperparameter tuning. The nestedness adds an additional internal cross validation loop to guarantee clean distinction between the 'test data' for model selection and tuning and ultimately the 'test data' used for model performance assessment.

Usually the data splits are then named "train" – "test" – "(external) validation", however different nomenclature is frequently used.

While resampling techniques can mitigate overfitting, they can also lead to manual overfitting when too many hyperparameter choices are made in the process[25]. Another consideration to keep in mind is that whenever a random data split is selected, it is with the assumption that each split is representative of the full data set. This can become problematic in two cases: (1) When data is dependent, data leakage occurs when train and test data share non-independent observations, such as the inclusion of both the index and revision surgery of patients. Both observations are systematically similar, induce overfitting and ultimately undermine the validity of the resulting model performance. (2) When data is not identically distributed: this is a serious problem in small sample scenarios, where splits are drawn out of a highly variable set of observations. Depending on which of the patients end up in the train or test data, the model performance can greatly fluctuate, and can be an overly optimistic estimate of predictive performance. Generally, less inflated predictive performance can be observed as the sample size increases [32]. As studies based on small sample sizes can generate highly variable estimates, conclusions may often be exaggerated or even invalid. Hence, predictive modeling should be restricted or used with caution when only small amounts of data are available. Considerations regarding sample size are discussed in *Part III*.

## Importance of External Validation in Clinical Prediction Modeling

External validation of clinical prediction models represents an important part in their development and rollout [5, 31]. In order to generalize, the input data, i.e. the training sample, needs to be *representative*. However, without external validation, the *site bias* or *center bias*, which includes variations in treatment protocols, surgical techniques, level of experience between departments and clinical users, as well as the so-called *sampling/selection bias*, which refers to systematically different data collection in regard to the patient cohort, cannot be detected. For these reasons, an empirical assessment of model performance on an unrelated, "external" dataset is required before an application can publicly be released. Erroneous or biased predictions can have severe sequelae for patients and clinicians alike, if misjudgments are made based upon such predictions. As a gold standard, external validation enables *unbiased testing* of model performance in a new cohort with different demographics. If a clinical prediction model shows comparable discrimination and calibration performance at external validation, generalizability may be confirmed. Then, it may be safe to release the model into the clinical decision-making progress. As an alternative to external validation – certainly the gold standard to ensure generalizability of a clinical prediction model – one might consider prospective internal validation (i.e. validation on a totally new sample of patients who are however derived from the same center with the same demographics, surgeons, and treatment protocols as the originally developed model). While prospective internal validation will also identify any overfitting that might be present, and will enable safe use of the prediction model at that specific center, this method does not allow ruling out center bias, i.e. does not ensure the safe use of the model in other populations.

## Feature Reduction and Selection

In overtly complex and high dimensional data with too many parameters, we find ourselves in an over-parameterized analytical setting. However, due 'the curse of dimensionality' - a term famously coined by Richard Bellmann in 1961 - generalization becomes increasingly more difficult in high dimensions. The approach to avoid "the curse" has been to find lower representation of the given feature space [4]. If there were too many features or variables present, *feature reduction* or *feature selection* methods can be applied. In *feature reduction*, methods are applied to simplify the complexity of the given high-dimensional data while retaining important and innate patterns of the data. Principal component analysis (PCA) is a popular illustration[24]. As an unsupervised ML method PCA is conceptually similar to clustering, and learns from data without any reference or a-priori knowledge of the predicted outcome. Analytically, PCA reduces high-dimensional data by projecting them onto the so-called principal components, which represent summaries of the data in fewer dimensions. PCA can hence be used as a strong statistical tool to reduce the main axis of variance within a given feature space. *Feature selection* refers to a similar procedure, which is also applied to initially too large feature spaces to reduce the number of input features. The key in feature selection is not to summarize data into lower dimensions as in feature reduction, but to actually reduce the number of included features to end up with only the "most useful" ones – and eliminate all non-informative ones. Naturally, if certain domain knowledge is present, vast sets of features can be constructed to a better set of informative features. For instance, in brain imaging, voxels of an MRI scan can either be considered individually or can be summarized into functionally or anatomically homogenous areas – a concept of topographical segregation that dates back to Brodmann[1, 11]. The problem of feature selection is well-known in the ML community and has generated a vast body of literature early on [3, 21]. A common pruning technique to select features that together maximize e.g. classification performance is *recursive feature elimination* (RFE)[13, 18]. In RFE, a given classifier or regressor is iteratively trained, and a ranking criterion for all features is estimated. The feature with the smallest respective ranking criterion is then eliminated. Introduced by Guyon and colleagues [13], RFE was initially used to extract small subsets of highly discriminant genes in DNA arrays and build reliable cancer classifiers. As an instance of backward elimination – that is, we start with the complete set of variables and progressively eliminate the least informative features – RFE can be used both in classification and regression settings with any given learner, but remains computationally greedy ("brute force"), as many different e.g. classifiers on feature subsets of decreasing size are revisited. As an important consideration, RFE selects *subsets* of variables based on an optimal *subset* ranking criterion. Consequently, a group of features combined may lead to optimal predictive performance, while the individual features included do not necessarily have to be the most important. Embedded in the process of model training, variable selection procedures such as RFE can improve performance by selecting subsets of variables that together maximize predictive power. Importantly, resampling methods should be applied when using RFE to factor in the variability caused by feature selection when calculating performance.

## Conclusion

Overfitting is a multifactorial problem, and there are just as many possible approaches to reduce its negative impact. We encourage the use of resampling methods such as cross-validation in every predictive modelling pipeline. While there are various options to choose from, we recommend the usage of k-fold cross validation or the bootstrap. Nested loops may be used for hyperparameter tuning and model selection. While the use of resampling does not solve overfitting, it helps to gain a more representative understanding of the predictive performance, especially of out-of-sample error. Feature reduction and selection methods, such as PCA and RFE are introduced for handling high-dimensional data. A potential pitfall along the way is *data contamination*, which occurs when data leaks from the resampled test to train set and hence leads to overconfident model performance. We encourage the use of standardized pipelines (see Part IV and V for examples), which include feature engineering, hyperparameter tuning and model selection within one loop to minimize the risk of unintentionally leaking test data. Finally, we recommend including a regularization term as hyperparameter and to restrict extensive model complexity, which will avoid overfitted predictive performance.

## Disclosures

Funding: No funding was received for this research.

Conflict of interest: All authors certify that they have no affiliations with or involvement in any organization or entity with any financial interest (such as honoraria; educational grants; participation in speakers' bureaus; membership, employment, consultancies, stock ownership, or other equity interest; and expert testimony or patent-licensing arrangements), or non-financial interest (such as personal or professional relationships, affiliations, knowledge or beliefs) in the subject matter or materials discussed in this manuscript.

Ethical approval: All procedures performed in studies involving human participants were in accordance with the ethical standards of the 1964 Helsinki declaration and its later amendments or comparable ethical standards.

Informed consent: No human or animal participants were included in this study.

# Machine learning-based clinical prediction modeling
## Part III: Evaluation and points of significance

*Julius M. Kernbach[1], *MD*; Victor E. Staartjes[2], *BMed*

[1] Neurosurgical Artificial Intelligence Laboratory Aachen (NAILA), Department of Neurosurgery,
RWTH Aachen University Hospital, Aachen, Germany
[2] Machine Intelligence in Clinical Neuroscience (MICN) Laboratory, Department of Neurosurgery,
University Hospital Zurich, Clinical Neuroscience Center, University of Zurich, Zurich, Switzerland

Corresponding Author
Victor E. Staartjes, BMed
Machine Intelligence in Clinical Neuroscience (MICN) Laboratory
Department of Neurosurgery, University Hospital Zurich
Clinical Neuroscience Center, University of Zurich
Frauenklinikstrasse 10
8091 Zurich, Switzerland
Tel +41 44 255 2660
Fax +41 44 255 4505
Web www.micnlab.com
E-Mail victoregon.staartjes@usz.ch

* J.M. Kernbach and V.E. Staartjes have contributed equally to this series, and share first authorship.

Abstract

In this section, we touch on the evaluation of clinical classification and regression models, as well as on which metrics should be reported when publishing predictive models. First, various available metrics to describe model performance in terms of discrimination (area under the curve (AUC), accuracy, sensitivity, specificity, positive predictive value, negative predictive value, F1 Score) and calibration (slope, intercept, Brier score, expected/observed ratio, Estimated Calibration Index, Hosmer-Lemeshow goodness-of-fit) are presented. The concept of recalibration is introduced, with Platt scaling and Isotonic regression as proposed methods. We discuss several of the common caveats and other points of significance that must be considered when developing a clinical prediction model. As such, we discuss considerations regarding the sample size required for optimal training of clinical prediction models - explaining why low sample sizes lead to unstable models, and offering the common rule of thumb of at least 10 patients per class per input feature, as well as some more nuanced approaches. Missing data treatment and model-based imputation instead of mean, mode, or median imputation is also discussed. We explain how data standardization is important in pre-processing, and how it can be achieved using e.g. centering and scaling. One-hot encoding is discussed - Categorical features with more than two levels must be encoded as multiple features to avoid wrong assumptions. Regarding binary classification models, we discuss how to select a sensible predicted probability cutoff for binary classification using the closest-to-(0,1)-criterion based on AUC, or based on the clinical question (rule-in or rule-out). Extrapolation is also discussed.

## Introduction

Once a dataset has been adequately prepared and a training structure (e.g. with a resampling method such as *k*-fold cross validation, c.f. *Part II*) has been set up, a model is ready to be trained. Already during training and the subsequent model tuning and selection, metrics to evaluate model performance become of central importance, as the hyperparameters and parameters of the models are tuned according to one or multiple of these performance metrics. In addition, after a final model has been selected based on these metrics, internal or external validation should be carried out to assess whether the same performance metrics can be achieved as during training. This section walks the reader through some of the common performance metrics to evaluate the discrimination and calibration of clinical prediction models based on machine learning (ML). We focus on clinical prediction models for continuous and binary endpoints, as these are by far the most common clinical applications of ML in neurosurgery. Multiclass classification – thus, the prediction of a categorical endpoint with more than two levels – may require other performance metrics.

Second, when developing a new clinical prediction model, there are several caveats and other points of significance that the readers should be aware of. These include what sample size is necessary for a robust model, how to pre-process data correctly, how to handle missing data and class imbalance, how to choose a cutoff for binary classification, and why extrapolation is problematic. In the second part of this section, these topics are sequentially discussed.

## Evaluation of Classification Models

### The Importance of Discrimination and Calibration

The performance of classification models can roughly be judged along two dimensions: Model discrimination and calibration.[22] The term *discrimination* denotes the ability of a prediction model to correctly classify whether a certain patient is going to or is not going to experience a certain outcome. Thus, discrimination described the accuracy of a binary prediction – yes or no. *Calibration*, however, describes the degree to which a model's predicted probabilities (ranging from 0% to 100%) correspond to the actually observed incidence of the binary endpoint (true posterior). Many publications do not report calibration metrics, although these are of central importance, as a well-calibrated predicted probability (e.g. your predicted probability of experiencing a complication is 18%) is often much more valuable to clinicians – and patients! – than a binary prediction (e.g. you are likely not going to experience a complication).[22]

There are other factors that should be considered when selecting models, such as complexity and interpretability of the algorithm, how well a model calibrates out-of-the-box, as well as e.g. the computing power necessary.[7] For instance, choosing an overly complex algorithm for relatively simple data (i.e. a deep neural network for tabulated medical data) will vastly increase the likelihood of overfitting with only negligible benefits in performance. Similarly, even though discrimination performance may be ever so slightly better with a more complex model such as a neural network, this comes at the cost of reduced interpretability ("black box" models).[21] The term "black box" model denotes a model for which we may know the input variables are fed into it and the predicted outcome, although there is no information on the inner workings of the model, i.e. why a certain prediction was made for an individual patient and which variables were most impactful. This is often the case for highly complex models such as deep neural networks or gradient boosting machines. For these models, usually only a broad "variable importance" metric that described a ranking of the input variables in order of importance can be calculated and should in fact be reported. However, how exactly the model integrated these inputs and arrived at the prediction cannot be comprehended in highly complex models.[21] In contrast, simpler ML algorithms, such as generalized linear models (GLMs) or generalized additive models (GAMs), which often suffice for clinical prediction modeling, provide interpretability in the form of odds ratios or partial dependence metrics, respectively. Lastly, highly complex models often exhibit poorer calibration out-of-the-box.[7]

Consequently, the single final model to be internally or externally validated, published, and readied for clinical use should not only be chosen based on resampled training performance.[23] Instead, the complexity of the dataset (i.e. tabulated patient data versus a set of DICOM images) should be taken into account. Whenever suitable, highly interpretable models such as generalized linear models or generalized additive models should be used. Overly complex models such as deep neural networks should generally be avoided for basic clinical prediction modelling.

## Model Discrimination

For a comprehensive assessment of model discrimination, the following data are necessary for each patient in the sample: A true outcome (also called "label" or "true posterior"), the predicted probabilities produced by the model, and the classification result based on that predicted probability (predicted outcome). To compare the predicted outcomes and the true outcomes, a confusion matrix (Table 1) can be generated. Nearly all discrimination metrics can then be derived from the confusion matrix.

### Area Under The Curve (AUC)

The only common discrimination metric that cannot be derived directly from the confusion matrix is the area under the receiver operating characteristic curve (AUROC, commonly abbreviated to AUC or ROC, also called *c*-statistic). For AUC, the predicted probabilities are instead contrasted with the true outcomes. The curve (Figure 1) shows the performance of a classification model at all binary classification cutoffs, plotting the true positive rate (Sensitivity) against the false positive rate (1 - Specificity). Lowering the binary classification cutoff classifies more patients as positive, thus increasing both false positives and true positives. It follows that AUC is the only common discrimination metric that is uniquely not contingent upon the chosen binary classification cutoff. The binary classification cutoff at the top left point of the curve, known as the "closest-to-(0,1)-criterion", can even be used to derive an optimal binary classification cutoff, which is explained in more detail further on.[15] Models are often trained and selected for AUC, as AUC can give a relatively broad view of a model's discriminative ability. An AUC value of 1.0 indicates perfect discrimination, while an AUC of 0.5 indicates a discriminative performance not superior to random prediction. Usually, a model is considered to perform well if an AUC of 0.7 or 0.8 is achieved. An AUC above 0.9 indicated excellent performance.

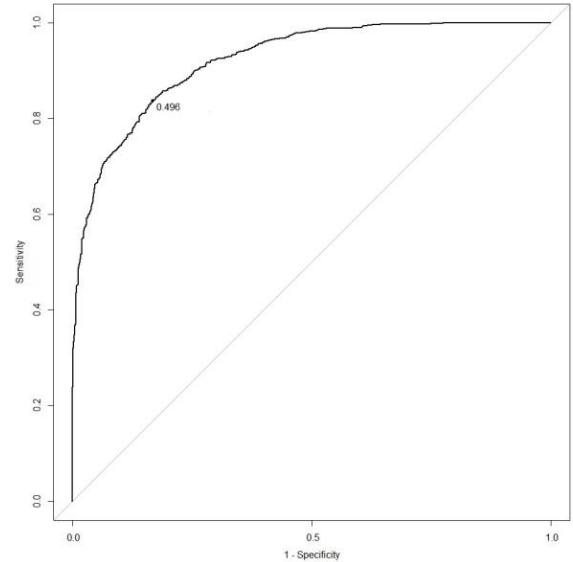

Figure 1 Area under the receiver operating characteristic curve (AUC) plot demonstrating an AUC of 0.922. The plot also indicated that, according to the "closest-to-(0,1)-criterion", 0.496 is the optimal binary classification cutoff that balances sensitivity and specificity perfectly.

### Accuracy

$$Accuracy = \frac{TP + TN}{P + N}$$

Based on the confusion matrix, a model's accuracy equals the total proportion of patients who were correctly classified as either positive or negative cases. While accuracy can give a broad overview of model performance, it is important to also consider sensitivity and specificity, as accuracy can be easily skewed by several factors including class imbalance (a caveat discussed in detail later on). An accuracy of 100% is optimal, while an accuracy of 50% indicates a performance that is equal to random predictions. The confusion matrix in Table 1 gives an accuracy of 83.5%.

### Sensitivity and Specificity

$$Sensitivity = \frac{TP}{P}$$
$$Specificity = \frac{TN}{N}$$

Sensitivity denotes the proportion of patients who are positive cases and who were indeed correctly predicted to be positive. Conversely, specificity measures the proportion of patients who are negative cases, and who were correctly predicted to be negative. Thus, a prediction model with high sensitivity generates only few false negatives, and the model can be used to "rule out" patients if the prediction is negative. A model with high specificity, however, can be used to "rule in" patients if positive, because it produces only few false positives. In data science, sensitivity is sometimes called "recall". The confusion matrix in Table 1 gives a sensitivity of 83.3% and a specificity of 83.6%.

|  | Negative Label | Positive Label |
|---|---|---|
| Predicted Negative | 800 (*True Negative*) | 174 (*False Negative*) |
| Predicted Positive | 157 (*False Positive*) | 869 (*True Positive*) |

Table 1 A confusion matrix

*Positive Predictive Value (PPV) and Negative Predictive Value (NPV)*

$$PPV = \frac{TP}{TP + FP}$$
$$NPV = \frac{TN}{TN + FN}$$

PPV is defined as the proportion of positively predicted patients who are indeed true positive cases. Conversely, NPV is defined as the proportion of negatively predicted patients who turn out to be true negatives. PPV and NPV are often said to be more easily clinically interpretably in the context of clinical prediction modeling than sensitivity and specificity, as they relate more directly to the prediction itself: For a model with a high PPV, a positive prediction is very likely to be correct, and for a model with a high NPV, a negative prediction is very likely to be a true negative. In data science, PPV is sometimes called "precision". The confusion matrix in Table 1 gives a PPV of 84.7% and a NPV of 82.1%.

*F1 Score*

$$F1 = 2 \times \frac{PPV \times Sensitivity}{PPV + Sensitivity}$$

The F1 score is a composite metric popular in the ML community, which is mathematically defined as the harmonic mean of PPV and sensitivity. Higher values represent better performance, with a maximum of 1.0. The F1 score is also commonly used to train and select models during training. The confusion matrix in Table 1 gives a F1 score of 0.840.

## Model Calibration

*Calibration Intercept and Slope*

As stated above, calibration describes the degree to which a model's predicted probabilities (ranging from 0%% to 100%) correspond to the actually observed incidence of the binary endpoint (true posterior). Especially for clinically applied models, a well-calibrated predicted probability (e.g. your predicted probability of experiencing a complication is 18%) is often much more valuable to clinicians and patients alike than a binary prediction (e.g. you are likely not going to experience a complication).[22] A quick overview of a model's calibration can be gained from generating a calibration plot (Figure 2), which we recommend to include for every published clinical prediction model. In a calibration plot, the patients of a certain cohort are stratified into $g$ equally-sized groups ranked according to their predicted probabilities. If you have a large cohort available, opt for $g = 10$; if you have only few patients you may opt for $g = 5$ to smooth the calibration curve to a certain degree. On the y axis, for each of the $g$ groups, the observed proportion of positive cases is plotted, while the mean predicted probability for each group is plotted on the x axis. A model with perfect calibration will have a calibration curve closely resembling a diagonal line. A poorly calibrated model will deviate in some way from the ideal diagonal line, or simply show an erratic form. From the predicted probabilities and the true posteriors, the two major calibration metrics can be derived: Calibration intercept and slope.[25]

The calibration intercept, also called "calibration-in-the-large", is a measure of overall calibration – A perfectly calibrated model has an intercept of 0.00. A model with a calibration intercept much larger than 0 generally puts out too high predicted probabilities –and thus overestimates the likelihood of a positive outcome. Likewise, a model with a negative intercept systematically underestimates probabilities. The model depicted in Figure 1 sports an intercept of 0.04.

The calibration slope quantifies the increase of true risk compared to predicted risk. A perfectly calibrated model has an intercept of 1.00. If a model has a calibration slope that is much larger than 1, the increase of the predicted probabilities on the calibration curve is too steep, and vice versa.

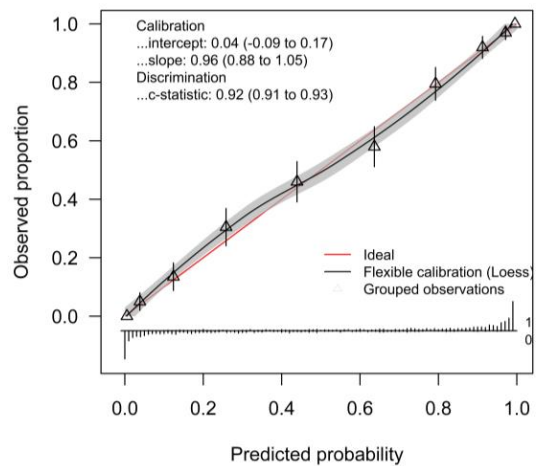

Figure 2 Calibration plot comparing the predicted probabilities – divided into ten bins - of a binary classification model to the true observed outcome proportions. The diagonal line represents the ideal calibration curve. A smoother has been fit over the ten bins. This model achieved an excellent calibration intercept of 0.04, with a slope of 0.96.

*Brier Score*

The Brier score [3] measures overall calibration and is defined as the average squared difference between predicted probabilities and true outcomes. It takes on values between 0 and 1, with lower values indicating better calibration. As a proper scoring rule, the Brier score simultaneously captures calibration itself as well as sharpness: A property that measures how much variation there is in the true probability across predictions. When assessing the performance of different binary classification models, the Brier score is mainly used to compare model performances, and – being mainly a relative measure – the actual value of the score is only of limited value. As a caveat, the Brier score only inaccurately measures calibration for rare outcomes.

*Other Calibration Metrics*

Various other calibration metrics have been developed, of which the following three are more commonly used. First, the expected/observed ratio, or E/O-ratio, describes the overall calibration of a prediction model, and is defined as the ratio of expected positive (predicted positive) cases and observed positive (true positive) cases.[16] A value of 1 is optimal. Second, the Estimated Calibration Index (ECI) [27] is a measure of overall calibration, and is defined as the average squared difference of the predicted probabilities with their grouped estimated observed probabilities. It can range between 0 and 100, with lower values representing better overall calibration. Lastly, the Hosmer-Lemeshow goodness-of-fit test can be applied to assess calibration, and is based on dividing the sample up according to $g$ groups of predicted probabilities, with $g = 10$ being a common value.[8] The test then compares the distribution to a chi-square distribution. A $p > 0.2$ is usually seen as an indication of a good fit i.e. fair calibration.

*Recalibration Techniques*

Should you have arrived at a robustly validated model with high performance in discrimination but poor calibration, there are several methods available to recalibrate the model to fit a population with a knowingly different incidence of the endpoint, or to even out a consistent deformation of the calibration curve.[22] These scenarios are explained in some more detail below. Also, if a study reports development of a model as well as external validation of that model in a different population for which the model is recalibrated, both the recalibrated as well as the uncalibrated performance of the model in the external validation cohort have to be reported, as the uncalibrated performance is the only representative and unbiased measure of generalizability available. In the first case, a model may have been developed in a certain country in which the incidence of a certain outcome is 10%. If other authors want to apply the same exact model in a different country with a known incidence of this outcome that is higher at e.g. 20%, the model will systematically underestimate predicted probabilities – and thus have a negative intercept, while maintaining a calibration slope of around 1.0. To adjust for the difference in outcome incidence, the intercept of the model can be updated to recalibrate the model.[9] In the second case, calibration curves may consistently show a sigmoid or other reproducible deviation from the ideal diagonal calibration curve. Two commonly applied methods to improve the calibration of the predicted probabilities are logistic regression and isotonic regression. Logistic regression can be used to train a wrapper model that learns to even out the deviation. This technique is called logistic recalibration of Platt scaling.[14] Second, isotonic regression can be applied to recalibrate the model.[14] Isotonic (also called monotonic) regression is a nonparametric technique that for fitting a free-form line (such as a calibration plot) to a series of reference values (such as a perfect diagonal line), under the constraints that the fitted line has to be monotonically increasing and must lie as close to the reference values as feasible.[14]

It is important to stress here that we recommend recalibration only in these two cases listed above: On the other hand, if the calibration curve is erratic or a deformation of the calibration curve (e.g. sigmoid deformation) is not consistent among resamples or validation cohorts, we do not recommend recalibration.

## Evaluation of Regression Models

For regression problems, performance can only be evaluated by comparing the predicted value and the true value directly. There are three major performance metrics that are used to evaluate the performance of regressors: First, root mean square error (RMSE), defined as the standard deviation of the differences between the predicted and true values (residuals), explains the distribution of the residuals around the perfect predictions. A perfect RMSE would be 0.00, with lower values indicating better performance. Similarly, mean absolute error (MAE) measures the difference among the predicted and the true values directly. Thus, a MAE of 0.00 would indicate no error at all, with deviations from 0 indicating overall over- or underestimation of values. Lastly, the $R^2$ value, defined as the square of the of the correlation coefficient among predicted and true values, discloses what proportion of the variation in the outcome is explained by the model. Consequently, a $R^2$ value of 1.0 would indicate perfect explanatory power, and 0.00 would indicate zero explanatory power of the model. For regression models, a quantile-quantile plot can optionally be included to illustrate the relationship among predicted and true values over the entire dataset (c.f. Part V).

## Points of Significance

### Choosing a Cutoff for Binary Classification

For binary classifiers which produce predicted probabilities, a cutoff (or threshold) has to be set to transform the predicted probabilities – ranging from 0.00 to 1.00 – to a binary classification (i.e. yes/no or positive/negative or 1/0). While it might be tempting and often adequate to simply always use a cutoff of 0.50 for binary classification, in many cases different cutoffs will produce more accurate results in general, and the cutoff should also be chosen depending on the intended application of a given model.

One quantitative method to calculate a cutoff for binary classification that optimizes both sensitivity and specificity is the AUC-based "closest-to-(0,1)-criterion" or Youden's index.[15] Using packages in R such as pROC [19], this can be done easily. This technique will lead to the most balanced estimation of a binary classification cutoff, and can be chosen on a model with generally high performance measures that is aimed at achieving maximum classification *accuracy* overall. However, in many cases, models are clinically intended to *rule-in* or *rule-out* critical events. In these cases, the binary classification cutoff may be adjusted to achieve high specificity or sensitivity, respectively. For example, a rule-in model requires a high specificity, whereas sensitivity is of secondary importance. In other words, if a model with high specificity (> 90%) makes a positive prediction, this rules in true case positivity, while a negative prediction will have rather little value. To increase specificity, the cutoff for binary classification can be adjusted *upwards* (e.g. to 75%). This can be seen as a "higher burden of proof" for a positive case. Inversely, a rule-out model will require high sensitivity and a negative result to rule out an event, in which case the cutoff for binary classification can be adjust *downwards*. Whether a clinical prediction model is laid out as a neutral model (cutoff 0.50 or calculated using the "closest-to-(0,1)-criterion"), rule-in model (cutoff adjusted upwards), or rule-in model (cutoff adjusted downwards) will depend on the clinical question.

Again, it is important to stress at this point that the selection of a cutoff for binary classification must occur using exclusively training data. Based on the (resampled) training performance, a cutoff should be chosen using one of the methods described above. Only one final, fully trained model and its determined cutoff should then be tested on the internal or external validation data, which will confirm the generalizability of both the model parameters and the cutoff that was chosen. If the cutoff is post-hoc adjusted based on the internal or external validation data, which are intended to provide an assessment of likely "real-world" performance, this evaluation of generalizability becomes rather meaningless and generalizability cannot be assessed in an unbiased way. Lastly, the threshold for binary classification should be reported when publishing a clinical prediction model.

### Sample Size

While even the largest cohort with millions of patients is not guaranteed to result in a robust clinical prediction model if no relevant input variables are included ("garbage in, garbage out" – do not expect to predict the future from age, gender, and body mass index), the relationship among predictive performance and sample size is certainly directly proportional, especially for some data-hungry ML algorithms. To ensure generalizability of the clinical prediction model, the sample size should be both representative enough of the patient population, and should take the complexity of the algorithm into account. For instance, a deep neural network – as an example of a highly complex model – will often require thousands of patients to converge, while a logistic regression model may achieve stable results with only a few hundreds of patients. In addition, the number of input variables plays a role. Roughly, it can be said that a bare minimum of 10 positive cases are required per included input variable to model the relationships. Often, erratic behavior of the models and high variance in performance among splits is observed when sample sizes are smaller than calculated with this rule of thumb. Of central importance is also the proportion of patients who experience the outcome. Also, for very rare events, a much larger total sample size is consequentially needed. For instance, a prediction based on 10 input features for an outcome occurring in only 10% of cases would require at least 1000 patients including at least 100 who experienced the outcome, according to the above rule of thumb. In general and from personal experience, we do not recommend developing ML models on cohorts with less than 100 positive cases and reasonably more cases in total, regardless of the rarity of the outcome. Also, one might consider the available literature on risk factors for the outcome of interest: If epidemiological studies find only weak associations with the outcome, it is likely that one will require more patients to arrive at a model with good predictive performance, as opposed to an outcome which has several highly associated risk factors, which may be easier to predict. Larger sample sizes also allow for more generous evaluation through a larger amount of patient data dedicated to training or validation, and usually results in better calibration measures. Lastly, some more nuanced and protocolized methods to arrive at a sample size have been published, such the Riley et al. expert's consensus on deriving a minimum sample size for generating clinical prediction models, which can also be consulted.[17, 18]

## Standardization

In clinical predictive modeling, the overall goal is to get the best discriminative performance from your ML algorithm, and some small steps to optimize your data before training may help to increase performance. In general, ML algorithms benefit from standardization of data, as they may perform more poorly if individual features to not appear more or less like normally distributed, e.g. representing Gaussian data with a mean value of 0 and a variance of 1. While most algorithms handle other distributions with ease, some (e.g. support vector machines with a radial basis function) assume centered and scaled data. If one input feature is orders of magnitude larger than all others, this feature may predominantly influence predictions and may decrease the algorithm's ability to learn from the other input data. In data science, centering and scaling are common methods of standardizing your data. To center continuous variables, means are subtracted from each value to arrive at a mean of 0. Scaling occurs through dividing all variables through their standard deviation, after which you end up with $z$ scores (the number of standard deviations a value is distanced from the mean). As an alternative to this *standardization* approach, data can also be *normalized*. This means that data are rescaled between their minimum and maximum (also called Min-Max-Scaling) to take on values from 0 to 1, which is particularly useful when data do not approximately follow a Gaussian distribution, in which case standardization based on standard deviations could lead to skewed results. Sometimes it can also be advantageous to transform i.e. logarithmically distributed variables. These steps are well-integrated into R through the caret package (c.f. *Parts IV and V*).[10] There are many other methods to pre-process data, which are also partially discussed in Part II and below. At this point, it is important to stress that all pre-processing steps should take place after data splitting into training and testing sets, as data leakage can occur (c.f. *Part II*).

## One-Hot Encoding

In many cases, dealing with categorical data as opposed to continuous data is challenging in data science. Especially when categorical variables are not ordinal, handling them similarly to continuous data can lead to wrong assumptions as to relationships among variables. In addition, some algorithms cannot work with categorical data directly and require numerical inputs instead, often rather due to their specific implementation in statistical programming languages like R and not due to a hard mathematical limitations of the algorithm itself. Take the variable "Histology", with the levels "Glioblastoma", "Low-grade Glioma", "Meningioma", and "Ependymoma" as an example of a non-ordinal feature. In the "Histology" variable, the encoding of the four levels in numerical form as "1" to "4" (simple *integer* encoding) would yield catastrophic results, as the four levels would be interpreted as a continuous variable and the level encoded as "4" is not necessarily graded higher as the level encoded as "1". This may lead to poorer performance and unanticipated results. In addition, any explanatory power, such as derived from measures of variables importance, will no longer be correct and would lead to clinical misinterpretation.

Instead, categorical variables with more than two levels should be *one-hot encoded*. This means that the original variable is removed, and that for each unique level of this categorical variable, a new dichotomous variable is created with the values "0" and "1". Thus, for the above "Histology" example, four new dichotomous variables will be created, namely "Glioblastoma [0,1]", "Low-grade Glioma [0,1]", and so forth. One-hot encoding ensures that the influence of each individual level of a categorical variable on the dependent variable can be accurately represented.

## Missing Data and Imputation

There are also other considerations in pre-processing other than centering and scaling, including the handling of missing data. In ideal circumstances, we would prefer to only work with complete datasets, but we are mostly faced with various amount of missing values. To deal with missing values is a science on its own, which has generated a vast body of literature [12, 20] and analytical strategies, broadly classified in either deletion ("complete case analysis") or imputation. In cases, in which values are missing at random (MAR) or completely at random (MCAR), it is safe to discard single observations with missing values or even complete feature columns when e.g. more than >50% of the column's observations are unaccounted for. When values are systematically missing instead, dropping features or observations subsequently introduces bias. In this case, imputation might yield better results. Strategies can range from simple approaches such as mean, mode, or median imputation, which, however, defeat the purpose of imputation for clinical prediction modeling since they do not factor in correlations between variables and do not work well with categorical variables, to more complex algorithmic imputation techniques. Any applied imputation method should however be used with care, and its necessity should be carefully considered especially when the fraction of missing data is substantial. The best approach will always be to keep missing data at a minimum.

There are also situations when imputing missing data may not be strictly necessary. First, some implementations of certain algorithms – for example the popular XGBoost [6] implementation of boosted decision trees – can handle missing data natively by treating empty fields as a unique value. However, the majority of algorithms will simply throw out any patients with missing data, or impute automatically. An additional point to consider is that, while some algorithms may be theoretically able to handle missing data natively, there is no reason that they should be made to do so. One of the cases in which imputing data is not strictly necessary is when an abundance of data is available with only few fields missing, or when data is missing for only a certain few patients – in which case it may be more convenient to simply delete the missing observations. Deleting larger amounts of data – as stated above – is not recommended because it may introduce systematic bias. More importantly, when data is clearly missing not at random (MNAR), imputation and deletion both will lead to inaccurate results, and the missingness must be explicitly modeled.[13] MNAR occurs when missingness depends on specific values of the data, e.g. when there is a systematic bias such as when certain populations are much less likely to return for follow-up visits due to geographical distance, or when obese patients are much less likely to report their weight. In cases with data that is MNAR, simple imputation will not yield correct results.

However, in the majority of cases it is advisable to co-train an imputer with the actual clinical prediction model, even if there is no missing data in the training set. This will allow for easy handling of missing data that the model may come across in the future, e.g. in an external validation set. Again, it is important to stress that as with all pre-processing steps, the co-trained imputer should only ever be trained on the training dataset of the prediction model – and should never see the validation data. Otherwise, data leakage may occur (c.f. Part II). Several simple packages for imputation exist in R, including algorithmic imputation using the *k*-nearest neighbour (kNN) algorithm.[26] In this approach, the missing datapoint is imputed by the average of *k* nearest neighbouring datapoints based on a chosen distance metric. In addition, single imputation can be achieved by simple regression models and predictive mean matching (PMM).[11] This approach works for both continuous and categorical variables, as a regressor predicts the missing value from the other available patient data, and then subsequently imputes the most closely matching value from the other patients without missing values. The advantage here is avoidance of imputation of extreme or unrealistic values, especially for categorical variables. This approach can also be extended to the state-of-the-art of multiple imputation through multivariate imputation based on chained equations (MICE) [4], which is harder to implement but depicts the uncertainty of the missing values more accurately.

While imputation can be achieved using many different algorithms including the methods described, we selected the nonparametric kNN method for internal consistency in both regression and classification (c.f. *Parts IV and V*) in our practical examples, and because there is some evidence that kNN-based imputation may outperform some other imputation methods.[1] In addition, kNN imputers are highly computationally efficient.

## Class Imbalance

Class imbalance is evident when one class - the minority class (i.e. patients who experienced a rare complication) - is much rarer than the majority class (i.e. patients who did not experience this rare complication).[24] In clinical neurosciences, class imbalance is a common caveat, and many published models do not adjust for it. Because ML models extract features better and are most robust if all classes are approximately equally distributed, it is important to know how to diagnose and counteract class imbalance. If a considerable amount of class imbalance is present, ML models will often become "lazy" in learning how to discriminate between classes and instead choose to simply vote for the majority class. This bias provides synthetically high AUC, accuracy, and specificity. However, sensitivity will be near zero, making the model unemployable. This "accuracy paradox" denotes the situation when synthetically high accuracy only reflects the underlying class distribution in unbalanced data. For instance, if sensitivity and specificity are not reported, class imbalance can still be spotted if the model accuracy is virtually identical to the incidence of the majority class. In general, if class imbalance is present, care should be taken to weight classes or to under- or oversample using data science techniques. Accuracy and AUC alone do not always give a full representation of an ML model's performance. This is

why reporting a minimum of sensitivity and specificity is crucial.[24]

As an example, one might want to predict complications from a cohort containing 90% of patients without complications. By largely voting for the majority class (no complication), the model would achieve an accuracy and specificity of around 90% and very low sensitivity without actually learning from the data. This can be countered by adjusting class weights within the model, by undersampling and thus removing observations from the majority class, or by oversampling the minority class.[2] Specifically, the synthetic minority oversampling technique (SMOTE) has been validated, shows robust performance, and is easy to employ.[5] SMOTE simulates new observations for the minority class by using *k*-means clustering, thus generating "synthetic" patients that have realistic characteristics derived from similar patients already present in the dataset. However, conventional upsampling – the simple copying of randomly selected patients of the minority class until class balance is achieved – often works similarly well. When training models, the method of handling class imbalance (i.e. none, conventional upsampling, or SMOTE) may be regarded as a hyperparameter

## Extrapolation

The vast majority of ML models are only capable of interpolating data – thus, making predictions on cases similar to the ones available in the training data – and are incapable of extrapolating – making predictions on situations that are relevantly different. This can be seen similarly to trying to apply the results of a randomized controlled drug trial to patients who were excluded from the study. For example, a model that predicts neurological impairment after brain tumor surgery that has been trained and externally validated on a large cohort of patients from 30 to 90 years of age should not be expected to make accurate predictions for pediatric brain tumor patients. Although the goal when developing algorithms is generalization to slightly other demographics, most algorithms learn to fit the training data as closely as possible locally, regardless of potential other situations not included in the training dataset (c.f. *Part II*). Thus, caution must be taken when making predictions outside the bounds of the type of patients included in the training data. Some algorithms are considered as being more prone to extrapolation errors, such as GAMs based on locally estimated scatterplot smoothing (LOESS) due to their reliance on local regression. In conclusion, trained models should not be clinically expected to extrapolate to patients with vastly differing characteristics.

Conclusion

Various metrics are available to evaluate the performance of clinical prediction models. A suggested minimum set of performance metrics includes AUC, accuracy, sensitivity, specificity, PPV, and NPV along with calibration slope and intercept for classification models, or RMSE, MAE, and $R^2$ for regression models. These performance metrics can be supplemented by a calibration plot or a quantile-quantile plot, respectively. Furthermore, there are some common caveats when developing clinical prediction models that readers should be aware of: Sample sizes must be sufficiently large to allow for adequate extraction of generalizable interactions among input variables and outcome and to allow for suitable model training and validation. Class imbalance has to be recognized and adjusted for. Missing data has to be reported and, if necessary, imputed using state-of-the-art methods. Trained models should not be clinically expected to extrapolate to patients with vastly differing characteristics. Finally, in binary classification problems, the cutoff to transform the predicted probabilities into a dichotomous outcome should be reported and set according to the goal of the clinical prediction model.

Disclosures

Funding: No funding was received for this research.
Conflict of interest: All authors certify that they have no affiliations with or involvement in any organization or entity with any financial interest (such as honoraria; educational grants; participation in speakers' bureaus; membership, employment, consultancies, stock ownership, or other equity interest; and expert testimony or patent-licensing arrangements), or non-financial interest (such as personal or professional relationships, affiliations, knowledge or beliefs) in the subject matter or materials discussed in this manuscript.
Ethical approval: All procedures performed in studies involving human participants were in accordance with the ethical standards of the 1964 Helsinki declaration and its later amendments or comparable ethical standards.
Informed consent: No human or animal participants were included in this study.

# Machine learning-based clinical prediction modeling
## Part IV: A practical approach to binary classification problems

*Julius M. Kernbach[1], *MD*; Victor E. Staartjes[2], *BMed*

[1] Neurosurgical Artificial Intelligence Laboratory Aachen (NAILA), Department of Neurosurgery,
RWTH Aachen University Hospital, Aachen, Germany
[2] Machine Intelligence in Clinical Neuroscience (MICN) Laboratory, Department of Neurosurgery,
University Hospital Zurich, Clinical Neuroscience Center, University of Zurich, Zurich, Switzerland

Corresponding Author
Victor E. Staartjes, BMed
Machine Intelligence in Clinical Neuroscience (MICN) Laboratory
Department of Neurosurgery, University Hospital Zurich
Clinical Neuroscience Center, University of Zurich
Frauenklinikstrasse 10
8091 Zurich, Switzerland
Tel +41 44 255 2660
Fax +41 44 255 4505
Web www.micnlab.com
E-Mail victoregon.staartjes@usz.ch

* J.M. Kernbach and V.E. Staartjes have contributed equally to this series, and share first authorship.

Abstract

This section goes through all the steps required to train and validate a simple, machine learning-based clinical prediction model for any binary outcome, such as for example the occurrence of a complication, in the statistical programming language R. To illustrate the methods applied, we supply a simulated database of 10'000 glioblastoma patients who underwent microsurgery, and predict the occurrence of 12-month survival. We walk the reader through each step, including import, checking, and splitting of datasets. In terms of preprocessing, we focus on how to practically implement imputation using a k-nearest neighbor algorithm, and how to perform feature selection using recursive feature elimination. When it comes to training models, we apply the theory discussed in Parts I to III on a generalized linear model, a generalized additive model, a stochastic gradient boosting machine, a random forest, and a naïve Bayes classifier. We show how to implement bootstrapping and to evaluate and select models based on out-of-sample error. Specifically for classification, we discuss how to counteract class imbalance by using upsampling techniques. We discuss how the reporting of a minimum of accuracy, area under the curve (AUC), sensitivity, and specificity for discrimination, as well as slope and intercept for calibration - if possible alongside a calibration plot - is paramount. Finally, we explain how to arrive at a measure of variable importance using a universal, AUC-based method. We provide the full, structured code, as well as the complete glioblastoma survival database for the readers to download and execute in parallel to this section.

## Introduction

Predictive analytics are currently by far the most common application of machine learning in neurosurgery [3, 4, 23, 24], although the potential of machine learning techniques for other applications such as natural language processing, medical image classification, radiomic feature extraction, and many more should definitely not be understated.[7, 13, 17, 18, 22, 28, 30–32] The topic of predictive analytics also uniquely lends itself to introducing machine learning methods due to its relative ease of implementation. Thus, we chose to specifically focus on predictive analytics as the most popular application of machine learning in neurosurgery. This section of the series is intended to demonstrate the programming methods required to train and validate a simple, machine learning-based clinical prediction model for any binary endpoint. Prediction of continuous endpoints (regression) will be covered in Part V.

We focus on the statistical programming language R[16], as it is freely available and widely regarded as the state-of-the-art in biostatistical programming. Other programming languages such as Python are certainly equally suited to the kind of programming introduced here. While we elucidate all necessary aspects of the required code, a basic understanding of R is thus required. Basic R courses are offered at many universities around the world, as well as through numerous media online and in print. We highly recommend that users first make themselves familiar with the programming language before studying this section. Python is another programming language often used in machine learning. The same general principles and pipeline discussed here can be applied in Python to arrive at a prediction model.

At this point we again want to stress that this section is not intended to represent a single perfect method that will apply to every binary endpoint, and to every data situation. Instead, this section represents one possible, generalizable methodology, that incorporates most of the important aspects of machine learning and clinical prediction modelling.

To illustrate the methods applied, we supply a simulated database of 10'000 glioblastoma patients who underwent microsurgery, and predict the occurrence of 12-month survival. Table 1 provides an overview over the glioblastoma database. We walk the reader through each step, including import, checking, splitting, imputation, and pre-processing of the data, as well as variable selection, model selection and training, and lastly correct evaluation of discrimination and calibration. Proper visualization and reporting of machine learning-based clinical prediction models for binary endpoints are also discussed.

The centerpiece of this section is the provided R code (Supplement 1), which is intended to be used in combination with the provided glioblastoma database (Supplement 2). When executed correctly and in parallel with this section's contents, the code will output the same results as those achieved by the authors, which allows for immediate feedback. The R code itself is numbered in parallel to this section, and also contains abundant explanations which enable a greater understanding of the functions and concepts that are necessary to succeed in generating a robust model. Finally, the code is intended as a scaffold upon which readers can build their own clinical prediction models for binary classification, and can easily be modified to do so for any dataset with a binary outcome.

## 1.  Setup and Pre-Processing Data

### 1.1  R Setup and Package Installation

Installing the most recent version of R (available at https://cran.r-project.org/) as well as the RStudio graphic user interface (GUI) (available at https://rstudio.com/products/rstudio/download/) is recommended.[16] A core strength of the R programming language is its wide adoption, and thus the availability of thousands of high-end, freely downloadable software packages that facilitate everything from model training to plotting graphs. Running the "*pacman*"[19] codes in section 1.1 will automatically ensure that all packages necessary for execution of this code are installed and loaded into the internal memory. You will require an internet connection to download these packages. If you have a clean R installation and have no installed any of the packages yet, it might take multiple minutes to download and install all necessary data. The script also gives you the option to update your R installation, should you desire to do so.

### 1.2  Importing Data

Generally, it is easiest to prepare your spreadsheet in the following way for machine learning: First, ensure that all data fields are in numerical form. That is, both continuous and categorical variables are reported as numbers. We recommend formatting binary (dichotomous) categorical variables (i.e. male gender) as 0 and 1, and categorical variables with multiple levels (i.e. tumor type [Astrocytoma, Glioblastoma, Oligodendroglioma, etc.]) as 1, 2, 3, and so forth, instead of as strings (text). Second, we recommend always placing your endpoint of interest in the last column of your spreadsheet. The Glioblastoma dataset that is provided is already correctly formatted. To import the data from the Glioblastoma database in Microsoft Excel (Supplement 2), run the code in section 1.2. For R to find the Glioblastoma database on your computer, you can either store the .xlsx file in the same folder as the R script, or you have to manually enter the path to the .xlsx file, as demonstrated in Figure 1. You could also use RStudio's GUI to import the dataset.

### 1.3 Check the Imported Data
Run "str(df)" to get an overview of the imported data's structure. We see that all 22 variables are correctly imported, but that they are all currently handled by R as numerical ("num") variables, whereas some of them are categorical and should thus be handled in R as "factor" variables. An overview of the variables in the Glioblastoma database is provided in Table 1.

### 1.4 Reformat Categorical Variables
To reformat all categorical variables to "factor" variables in R, allocate them to the "cols" object. Subsequently, we apply the "factor" function to all columns of the database using the "lapply" function. Lastly, the binary endpoint of interest ("TwelveMonths") should be internally labelled as "yes" and "no", again using the "factor" function. Lastly, "str(df)" is used again to confirm the correct reformatting.

### 1.5 Remove Unnecessary Columns
Your imported data may contain extra columns with variables that are irrelevant to the current classification problem, such as patient numbers, names, or other endpoints. The latter is the case in the Glioblastoma database: The 21st column contains the variable "Survival", which is a continuous form of our binary endpoint of interest "TwelveMonths". Leaving this redundant variable in would lead to data leakage – the model could simply take the continuous "Survival" variable and extrapolate our endpoint "TwelveMonths" from it, without actually learning any features.

Columns can be removed from a R dataframe by specifying the column number to be removed. "Survival" is situated in the 21st column of the database, and can thus be removed by applying the function "df <- df[,-21]".

### 1.6 Enable Multicore Processing
If you are working on a machine with multiple central processing unit (CPU) cores, you can enable parallel computing for some functions in R. Using the code in section 1.6, create a computational cluster by entering the number of cores you want to invest into model development.[15] The default is set to 4, which is nowadays a common number of CPU cores.

### 1.7 Partition the Data for Training and Testing
Figure 2 illustrates the procedure. To randomly split the data into 80% for training an 20% for testing (internal validation), we first set a random seed, such as "set.seed(123)", although you could choose any number. Setting seeds initializes random functions in a constant way, and thus enables reproducibility. Subsequently, we randomly sample 80% of patients and allocate them to the training set ("train"), and do the same for the test set ("test"). Then, the rows of the two newly partitioned sets are shuffled, and the two sets are checked for an approximately equal distribution of the binary endpoint using the "prop.table()" function. Both sets contain around 52% of patients who survived for at least twelve months.

### 1.8 Impute Missing Data
The glioblastoma database contains no missing data, as the function "VIM::kNN"[29] will let you know. However, should you encounter missing data, this code block should automatically impute missing data using a *k*-nearest neighbor (KNN) algorithm.[1] It is important only to impute missing data within the training set, and to leave the test set alone. This is to prevent data leakage. Also, imputation can be achieved using many different algorithms. We elected to use a KNN imputer for reasons of consistency - during model training, a separate KNN imputer will be co-trained with the prediction model to impute any future missing data.

### 1.9 Variable Selection using Recursive Feature Elimination
Recursive Feature Elimination (RFE) is just one of various methods for variable selection (c.f. *Part II* for further explanation). In this example, we apply RFE (Figure 3) due to its relative simplicity, generalizability, and reproducibility. Because random functions are involved, seeds need to be set. A naive Bayes classifier is selected, and bootstrap resampling with 25 repetitions is used to ensure generalizability of the results. Suing the "sizes" argument in the "rfe" function[14], the number of combined variables that are to be assessed can be limited. As we have 20 independent variables, we choose to limit the search for the optimal number and combination of variables to between 10 and 20. The "rfe()" function is executed, which may take some minutes. Using "plot()", the results of the RFE procedure can be illustrated (Figure 4), and it is clear that a combination of 13 variables led to the highest performance. The selected variables are stored in "predictors(RFE)". These 13 selected variables, plus the endpoint "TwelveMonths" are stored in "keepvars", and the training set is subsequently reduced to 14 columns.

### 1.10 Get a Final Overview of the Data

Before diving directly into model training, it is advisable to look over the training and test set using the "summary()" function to assess the correctness of the independent variables and the endpoint.

## 2. Model Training

### 2.1 Setting up the Training Structure

Now that the data are prepared, training of the different models can be initiated. In this example, we elected to train five different algorithms to predict binary 12-month survival: Logistic regression[12] (generalized linear models, GLM), random forests[2] (RF), stochastic gradient boosting machines[9] (GBM), generalized additive models[11] (GAM), and naive Bayes classifiers[21] (NB). A brief overview of the five different models is provided in Table 2. We specifically refrained from using more complex models, such as neural networks, due to their inherently decreased interpretability and because they are more prone to overfitting on the relatively simple, clinical data used in this example.[8] All five models are trained sequentially and in a similar way using a universal wrapper that executes training, the "caret" package.[14] Hyperparameters – if available – are tuned automatically. To prevent overfitting, bootstrap resampling with 25 repetitions is chosen in this example (Figure 5).[25] However, 5-fold cross validation could also easily be implemented (c.f. Part V). To adjust for any potential class imbalance (c.f. Part III), random upsampling is implemented by choosing "sampling = "up"", although synthetic minority oversampling (SMOTE) could also be used (sampling = "smote").[5, 27] The current Glioblastoma dataset is however without class imbalance, as short-term and longer-term survivors are approximately equally common.

### 2.2 Model Training

The procedure (Figure 5) is equivalent for all five models (Sections 2.2.1 to 2.2.5). First, a seed is set to initialize the random number generator in a reproducible way. Subsequently, the algorithm to be used is specified in the "method" argument – the first model to be trained is a logistic GLM, so "method = "glm"" is chosen. The "tuneLength" argument depends on the complexity and of the hyperparameters: GLM has no hyperparameters, so a low value is specified. We specify that the parameters and hyperparameters are to be optimized according to area under the curve (AUC, metric = "ROC"), and that a KNN imputer is co-trained for future missing data (preProcess = "knnImpute"). The inputs are automatically centered and scaled by the "caret" package. After running the fully specified "caret::train" function, it may take some minutes for all resamples to finish training. The red "STOP" dot at the top right of the RStudio console will be present for as long as the model is training. Subsequently, a confusion matrix (conf) is generated, along with some other metrics that allow evaluation of the model's discrimination. Now, calibration is assessed and a calibration plot is generated using the "val.prob()" function.[10] Finally, the model specifications and resampled training performance are printed, and the model can be saved using the "save()" function for potential further use.

After completion of training the GLM (Section 2.2.1), the same procedure is repeated for the RF (Section 2.2.2), GBM (Section 2.2.3), GAM (Section 2.2.4), and NB (Section 2.2.5) models.

## 3. Model Evaluation and Selection

### 3.1 Model Training Evaluation

As soon as all five models have been trained, their performance on the training data can be compared. The final model should be selected based upon training data only. Criteria for clinical prediction model selection may include discrimination and calibration on the training set, as well as the degree of interpretability of the algorithm. Section 3.1 compiles the results of all five models, and allows their comparison in terms of discrimination (AUC, accuracy, sensitivity, specificity, positive predictive value (PPV), negative predictive value (NPV), F1 score) and calibration (intercept and slope) metrics.[26] The code in this section will also open a new plot viewer window using "dev.new()", that allows graphical comparison of the five models. If you have executed all parts of the script correctly up to this point, you will be presented with a plot that is identical to Figure 6. In this plot, we see that – while all models performed admirably – the GLM and GAM had the highest discrimination metrics. Models perform well if these discrimination measures approach 1. In addition, while all absolute values of intercept were very low, not all models had excellent calibration slopes. A perfectly calibrated model has an intercept of 0.0 and a slope of 1.0. Only the GLM and GAM had virtually perfect slopes. As both algorithms are highly interpretable, the GLM and the GAM both would make fine options for a final model. In this example, we elected to carry on with the GAM.

### 3.2  Select the Final Model
The fully trained GAM model was previously stored as "gamfit", and its resampled training evaluation as "GAM" in section 2.2.4. Now, as the GAM model is selected as the final model, it is renamed "finalmodel", and its training evaluation is renamed "finalmodelstats". You can choose any other model by replacing these two terms with the corresponding objects from section 2.2.

### 3.3  Internal Validation on the Test Set
For the first time since partitioning the original Glioblastoma database, the 20% of patients allocated to the test set are now used to internally validate the final model. First, a prediction is made on the test set using "finalmodel" and the "predict()" function. Of note, during prediction with the GAM on the test set, you will encounter warning messages indicating that extrapolation took place. These warning messages are not to be considered as errors, but as informative warnings indicating that some patients in the test set had characteristics that were outside of the bounds encountered by the GAM during training. GAMs rely on local regression, which makes extrapolation to extreme input values problematic. This is discussed in some more detail in Part III.

The predicted probabilities for the entire test set are then contrasted with the actual class labels from the endpoint (test$TwelveMonths) to arrive at an AUC value.[20] Subsequently, the predicted probabilities are converted into binary predictions using "factor(ifelse(prob$yes > 0.50, 'yes', 'no'))". Thus, predicted probabilities over 0.50 are counted as positive predictions (yes), and vice versa. This cutoff for binary classification can be changed to different values, changing sensitivity and specificity of the clinical prediction model. However, the decision to do so must be based solely on the training data, and thus already be taken before evaluation of the test set – otherwise, a clean assessment of out-of-sample error through internal validation is not possible anymore. This is discussed in some more detail in Part III. However, in most cases and especially with well-calibrated models, a standard cutoff of 0.50 is appropriate.

Subsequently, discrimination and calibration are calculated in the same way as previously. Using "print(Final)", the internal validation metrics can be viewed. Performance that is on par with or slightly worse than the training performance usually indicates a robust, generalizable model. Performance that is relevantly worse than the training performance indicates overfitting during training. These problems are discussed in detail in Part II. The final model can be saved, and will be available as "FINALMODEL.Rdata" in the same folder as the R script. Using the "load()" function, models can be imported back into R at a later date.

If you end up with the same performance metrics for the final GAM as in Table 3, you have executed all steps correctly.

## 4.  Reporting and Visualization
When generating clinical prediction models and publishing their results, there is a minimum set of information that ought to be provided to the reader. First, the training methods and exact algorithm type should be reported, if possible along with the code that was used for training. Second, the characteristics of the cohort that was used for training should be provided, such as in Table 1. If multiple cohorts are combined or used for external validation, the patient characteristics should be reported in separate. Discrimination and calibration must be reported. There are countless metrics to describe calibration and discrimination of prediction models. The bare minimum that should be reported for a binary prediction model probably consists of AUC, accuracy, sensitivity, specificity, PPV, and NPV, along with calibration intercept and slope. The F1 score can also be provided. A calibration plot should also be provided for binary prediction models.

Lastly, whenever feasible, an attempt at interpreting the model should be made. For example, logistic regression (GLM) models produce odds ratios, and GAMs can produce partial dependence values. However, there are also universal methods to generate variable importance measures that can apply to most binary prediction models, usually based on AUC, which we present below. To simplify reporting, this final section helps compile all these data required for publication of clinical prediction models. For further information on reporting standards, consult the transparent reporting of a multivariable prediction model for individual prognosis or diagnosis (TRIPOD) checklist.[6]

### 4.1 Compiling Training Performance
The resampled training performance in terms of discrimination and calibration can be printed using "print(finalmodelstats)". The metrics that are produced include AUC, accuracy, sensitivity (recall), specificity, PPV (precision), NPV, F1 score, intercept, and slope. Subsequently, a calibration plot is generated for the training set using the "val.prob()" function.

### 4.2 Compiling Internal Validation Performance
Similarly, the performance on the test set (internal validation) can be recapitulated, and a calibration plot produced (analogous to Figure 7).

### 4.3 Assessing Variable Importance
By using "varImp(finalmodel)", a universal method for estimation of variable importance based on AUC is executed, and results in a list of values ranging from 0 to 100, with 100 indicating the variable that contributed most strongly to the predictions, and vice versa. Finally, "plot(imp)" generates a variable importance plot that can also be included in publication of clinical prediction models (c.f. *Part V*).

### Conclusion
This section presents one possible and standardized way of developing clinical prediction models for binary endpoints. Proper visualization and reporting of machine learning-based clinical prediction models for binary endpoints are also discussed. We provide the full, structured code, as well as the complete Glioblastoma survival database for the readers to download and execute in parallel to this section. The methods presented can and are in fact intended to be extended by the readers to new datasets, new endpoints, and new algorithms.

### Disclosures
Funding: No funding was received for this research.

Conflict of interest: All authors certify that they have no affiliations with or involvement in any organization or entity with any financial interest (such as honoraria; educational grants; participation in speakers' bureaus; membership, employment, consultancies, stock ownership, or other equity interest; and expert testimony or patent-licensing arrangements), or non-financial interest (such as personal or professional relationships, affiliations, knowledge or beliefs) in the subject matter or materials discussed in this manuscript.

Ethical approval: All procedures performed in studies involving human participants were in accordance with the ethical standards of the 1964 Helsinki declaration and its later amendments or comparable ethical standards.

Informed consent: No human or animal participants were included in this study.

Figures

```
#A - Store the .xlsx file in the same folder as the script
df <- as.data.frame(read_excel("GlioblastomaData.xlsx"))

#B - Alternatively, you may have to enter the path to the .xlsx. file
df <- as.data.frame(read_excel("C:/Machine Learning/Folder/GlioblastomaData.xlsx"))
```

Figure 1 Code section 1.2: You can either import the Glioblastoma database by keeping its .xlsx file in the same folder as the R script (A). Alternatively, you may have to find the path (e.g. "C:/Desktop/database.xlsx") to the .xlsx file and enter it as a string, i.e. between quotes (B). Lastly, you could also use the graphic user interface of RStudio to import the file.

```
set.seed(123)    #Setting seeds allows reproducibility
dt = sort(sample(nrow(df), nrow(df)*.80)) #Allocate 80% split
train <-df[dt,] #Split
test <-df[-dt,]
train <- train[sample(nrow(train)),] #Shuffle
test <- test[sample(nrow(test)),]
```

Figure 2 Code section 1.7: This section illustrates how to partition a database into training and test (internal validation) sets.

```
rfecontrol <- rfeControl(functions=nbFuncs, method = "boot", number = 25, rerank = F, verbose = T, allowParallel = T)
set.seed(123)
RFE <- rfe(x = train[,1:(ncol(train)-1)], y = train[,ncol(train)], sizes=c(10:(ncol(train)-1)), rfecontrol=rfecontrol)
print(RFE) #Results
plot(RFE, type=c("g", "o")) #Plot performance over #vars
predictors(RFE) #variables to be kept
```

Figure 3 Code section 1.9: This section illustrates the recursive feature elimination (RFE) procedure. A naïve Bayes classifier is chosen, along with bootstrap resampling with 25 repetitions.

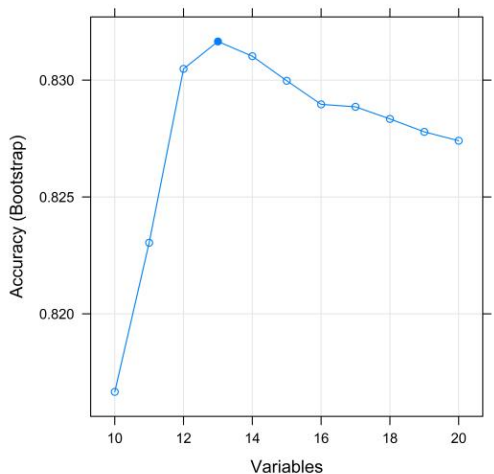

Figure 4 Results of the recursive feature elimination (RFE) variable election procedure. It was determined that using 13 variables explained the highest amount of variance, as seen in the superior accuracy that was achieved with this number and combination of variables.

```
#2.1 set up the Training and Resampling Structure ----
ctrl <- trainControl(method = "boot", number = 25, classProbs = T, summaryFunction = twoClassSummary, allowParallel = T, sampling = "up", returnResamp = "all")

#2.2.1 Train Model #1: GLM (Logistic Regression) ----
set.seed(123)
lrfit <- caret::train(TwelveMonths ~ ., data = train, method = "glm", trControl = ctrl, tuneLength = 5, metric = "ROC", preProcess = "knnImpute", na.action = na.pass)
#Assess Discrimination
conf <- print(confusionMatrix.train(lrfit, norm = "none", positive = "yes")[[1]][[1]]); tp <- conf[2,2]; tn <- conf[1,1]; fp <- conf[2,1]; fn <- conf[1,2];
sens <- tp/(tp+fn); spec <- tn/(tn+fp); prev <- as.numeric(prop.table(table(train[,ncol(train)])))[2]
#Assess Calibration
prob <- predict(lrfit, train, type="prob", na.action = na.pass)
calLogReg <- val.prob(prob[,2], as.numeric(train$TwelveMonths)-1, ylab = "Observed Frequency", xlab = "Predicted Probability", g = 10)
LogReg <- as.data.frame(cbind(max(lrfit$results$ROC), sens * prev + spec * (1-prev), sens, spec, (sens*prev)/(sens*prev+(1-spec)*(1-prev)),
        (spec*(1-prev))/((1-sens)*prev+spec*(1-prev)),
        (2*((sens*prev)/(sens*prev+(1-prev)))*sens))/(((sens*prev)/(sens*prev+(1-spec)*(1-prev)))+sens)))
LogReg <- as.data.frame(cbind(LogReg, calLogReg[12], calLogReg[13]))
colnames(LogReg)  <- c("AUC", "Accuracy", "Sensitivity", "Specificity", "PPV", "NPV", "F1 Score", "Intercept", "Slope")

#Summary
print(lrfit)
print(LogReg)
```

Figure 5 Code sections 2.1 & 2.2: First, the training structure is established: Bootstrap resampling with 25 repetitions is used. As a standard, random upsampling is applied to adjust for class imbalance if present. Subsequently, a logistic regression model (generalized linear model, GLM) is trained. All predictor variables are provided to the model, and it is automatically tuned for AUC. A k-nearest neighbor imputer is co-trained to impute any potential missing data in future predictions. Subsequently, discrimination and calibration are assessed, and the final model information and resampled training performance are printed.

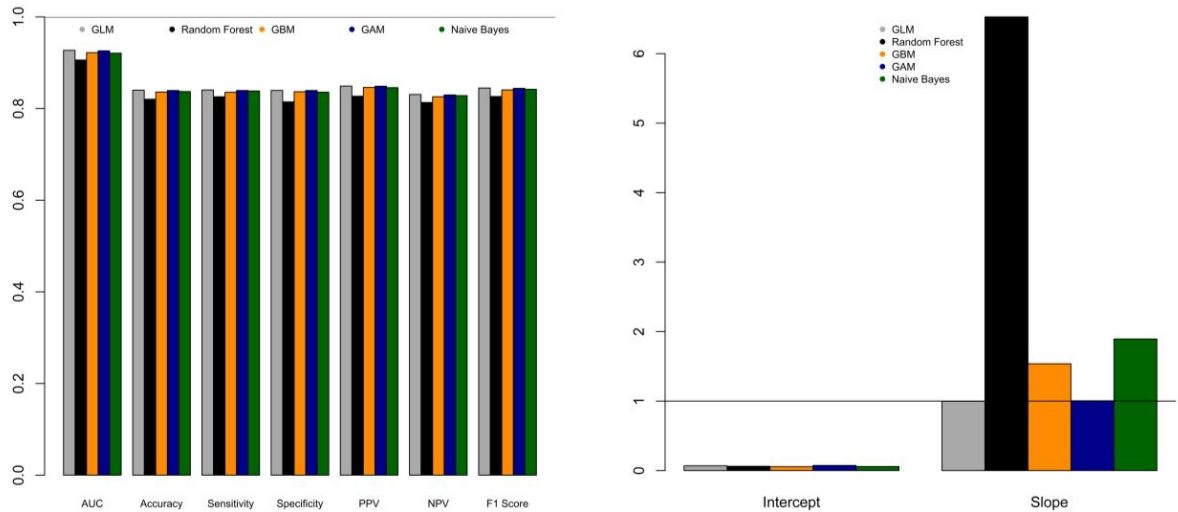

Figure 6 Graphical comparison of discrimination (left) and calibration (right) metrics (Code section 3.1). The GLM and GAM both exhibited the highest discrimination measures, with very low absolute intercept values and almost perfect slopes approaching 1.

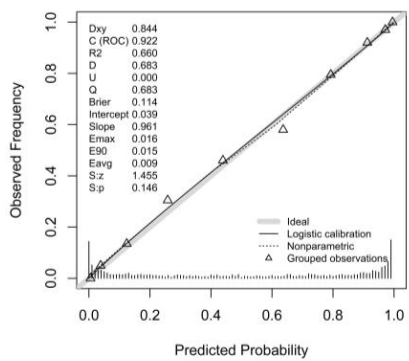

Figure 7 Calibration plot for the final GAM, demonstrating its calibration on the test set (internal validation). The calibration curve closely approximates the diagonal line, indicating excellent calibration.

Tables

Table 1 Structure of the simulated glioblastoma dataset. The number of included patients is 10'000. Values are provided as means and standard deviations or as numbers and percentages.

| Variable Name | Description | Value |
|---|---|---|
| Survival | Overall survival from diagnosis in months | 12.1 ± 3.1 |
| TwelveMonths | Patients who survived 12 months or more from diagnosis | 5184 (51.8%) |
| IDH | IDH mutation present | 4136 (41.4%) |
| MGMT | MGMT promoter methylated | 5622 (56.2%) |
| TERTp | TERTp mutation present | 5108 (51.1%) |
| Male | Male gender | 4866 (48.7%) |
| Midline | Extension of the tumor into the midline | 2601 (26.0%) |
| Comorbidity | Presence of any systemic comorbidity such as diabetes, coronary heart disease, chronic obstructive pulmonary disease, etc. | 5135 (51.4%) |
| Epilepsy | Occurrence of an epileptic seizure | 3311 (33.1%) |
| PriorSurgery | Presence of prior cranial surgery | 5283 (52.8%) |
| Married | Positive marriage status | 5475 (54.8%) |
| ActiveWorker | Patient is actively working, i.e. not retired, student, out of work, etc. | 5459 (54.6%) |
| Chemotherapy | Patients who received chemotherapy for glioblastoma | 4081 (40.8%) |
| HigherEducation | Patients who received some form of higher education | 4209 (42.1%) |
| Caseload | Yearly glioblastoma microsurgery caseload at the treating center | 165.0 ± 38.7 |
| Age | Patient age at diagnosis in years | 66.0 ± 6.2 |
| RadiotherapyDose | Total radiotherapy absorbed dose in Gray | 24.8 ± 6.7 |
| KPS | Karnofsky Performance Scale | 70.5 ± 8.0 |
| Income | Net yearly household income in US dollars | 268'052 ± 62'867 |
| Height | Patient body height in cm | 174.6 ± 6.7 |
| BMI | Deviation of body mass index from 25: in kg/m2 | 0.02 ± 1.0 |
| Size | Maximum tumor diameter in cm | 2.98 ± 0.55 |

Table 2 Overview of the five models that were employed.

| Model | caret::train() input | Package | Suitability | Hyperparameters |
|---|---|---|---|---|
| Generalized Linear Model | glm | stats | Classification, Regression | None |
| Random Forest | rf | randomForest | Classification, Regression | mtry (number of variables at each tree node) |
| Stochastic Gradient Boosting | gbm | gbm | Classification | n.trees (number of trees), interaction.depth (maximum nodes per tree), shrinkage (learning rate), n.minobsinnode (minimum number of patients per terminal node) |
| Generalized Additive Model | gamLoess | gam | Classification, Regression | span (smoothing span width), degree (degree of polynomial) |
| Naïve Bayes Classifier | nb | klaR | Classification | fL (Laplace correction factor), usekernel (normal or kernel density estimate), adjust |

Table 3 Performance metrics of the binary classification model (generalized additive model; GAM) for 12-month glioblastoma survival. The difference in performance among training and testing is minimal, demonstrating a lack of overfitting at internal validation.

| Metric | Cohort | |
|---|---|---|
| | Training (n = 8000) | Internal Validation (n = 2000) |
| Discrimination | | |
| AUC | 0.926 | 0.922 |
| Accuracy | 0.839 | 0.847 |
| Sensitivity | 0.839 | 0.848 |
| Specificity | 0.839 | 0.846 |
| PPV | 0.849 | 0.848 |
| NPV | 0.830 | 0.826 |
| F1 Score | 0.844 | 0.843 |
| Calibration | | |
| Intercept | 0.074 | 0.039 |
| Slope | 1.004 | 0.961 |

Supplementary Material

Please download the supplementary files from https://micnlab.com/files/

Supplement 1 R Code for binary classification of twelve-month survival of glioblastoma patients.

Supplement 2 Database of 10'000 simulated glioblastoma patients, intended for use with the R code

# Machine learning-based clinical prediction modeling
# Part V: A practical approach to regression problems

*Julius M. Kernbach[1], *MD*; Victor E. Staartjes[2], *BMed*

[1] Neurosurgical Artificial Intelligence Laboratory Aachen (NAILA), Department of Neurosurgery, RWTH Aachen University Hospital, Aachen, Germany
[2] Machine Intelligence in Clinical Neuroscience (MICN) Laboratory, Department of Neurosurgery, University Hospital Zurich, Clinical Neuroscience Center, University of Zurich, Zurich, Switzerland

Corresponding Author
Victor E. Staartjes, BMed
Machine Intelligence in Clinical Neuroscience (MICN) Laboratory
Department of Neurosurgery, University Hospital Zurich
Clinical Neuroscience Center, University of Zurich
Frauenklinikstrasse 10
8091 Zurich, Switzerland
Tel +41 44 255 2660
Fax +41 44 255 4505
Web www.micnlab.com
E-Mail victoregon.staartjes@usz.ch

* J.M. Kernbach and V.E. Staartjes have contributed equally to this series, and share first authorship.

Abstract

This section goes through the steps required to train and validate a simple, machine learning-based clinical prediction model for any continuous outcome, such as for example survival or a continuous outcome scale after surgery, in the statistical programming language R. To illustrate the methods applied, we supply a simulated database of 10'000 glioblastoma patients who underwent microsurgery, and predict survival from diagnosis in months. We walk the reader through each step, including import, checking, splitting of data. In terms of preprocessing, we focus on how to practically implement imputation using a k-nearest neighbor algorithm. We also illustrate how to select features based on recursive feature elimination, and how to use k-fold cross validation. When it comes to training models, we apply the theory discussed in Parts I to III on a generalized linear model, a generalized additive model, a random forest, a ridge regressor, and a Least Absolute Shrinkage and Selection Operator (LASSO) regressor. Specifically for regression, we discuss how to evaluate root mean square error (RMSE), mean average error (MAE), and the $R^2$ statistic, as well as how a quantile-quantile plot can be used to assess the performance of the regressor along the spectrum of the outcome variable, similarly to calibration when dealing with binary outcomes. Finally, we explain how to arrive at a measure of variable importance using a universal, nonparametric method. We provide the full, structured code, as well as the complete glioblastoma survival database for the readers to download and execute in parallel to this section.

## Introduction
In the neurosurgical literature, applications of machine learning for clinical prediction modeling are by far the most common [3, 4, 18, 19]. The topic of predictive analytics also uniquely lends itself to introducing machine learning methods due to its relative ease of implementation. Still, we chose to specifically focus on predictive analytics as the most popular application of machine learning in neurosurgery. Nonetheless, the great potential of machine learning methods in fields other than prediction modeling, such as e.g. natural language processing, medical image classification, radiomic feature extraction, and many more must not go unmentioned.[6, 10, 14, 15, 17, 22, 24–26] In clinical predictive analytics, those models concerned with prediction of continuous endpoints (e.g. survival in months) as opposed to binary endpoints (e.g. occurrence of a complication) are coined regressors. Regression problems, in contrast to classification problems, require different methodology, different algorithms, and different evaluation and reporting strategies.

Whereas Part IV laid out the details of generating binary prediction models, this section of the series is intended to demonstrate the programming methods required to train and validate a simple, machine learning-based clinical prediction model for any continuous endpoint. Many concepts and parts of the code have already been discussed in more detail in Part IV, and this part will focus on the differences to predicting binary endpoints. For a better understanding of the methods presented herein, Part IV should thus be studied first.

We focus on the statistical programming language R[13], as it is freely available and widely regarded as the state-of-the-art in biostatistical programming. While we elucidate all necessary aspects of the required code, a basic understanding of R is thus required. Basic R courses are offered at many universities around the world, as well as through numerous media online and in print. We highly recommend that users first make themselves familiar with the programming language before studying this section.

At this point we again want to stress that this section is not intended to represent a single perfect method that will apply to every continuous endpoint, and to every data situation. Instead, this section represents one possible, generalizable methodology, that incorporates most of the important aspects of machine learning and clinical prediction modelling.

To illustrate the methods applied, we supply a simulated database of 10'000 glioblastoma patients who underwent microsurgery, and predict the occurrence of 12-month survival. Table 1 provides an overview over the glioblastoma database. We walk the reader through each step, including import, checking, splitting, imputation, and pre-processing of the data, as well as variable selection, model selection and training, and lastly correct evaluation of the regression model. Proper reporting is also discussed.

The centerpiece of this section is the provided R code (Supplement 1), which is intended to be used in combination with the provided glioblastoma database (Supplement 2). When executed correctly and in parallel with this section's contents, the code will output the same results as those achieved by the authors, which allows for immediate feedback. The R code itself is numbered in parallel to this section, and also contains abundant explanations which enable a greater understanding of the functions and concepts that are necessary to succeed in generating a robust model. Finally, the code is intended as a scaffold upon which readers can build their own clinical prediction models for regression, and can easily be modified to do so for any dataset with a continuous outcome.

## 1.  Setup and Pre-Processing Data
Sections 1.1 to 1.3 are identical to those required for classification problems, and are thus covered in Part IV.[16] Thus, we kindly ask the reader to consult Part IV for further clarification on R setup, package loading, importing data, and checking the formatting of the imported data.

### 1.4 Reformat Categorical Variables
To reformat all categorical variables to "factor" variables in R, allocate them to the "cols" object. Subsequently, we apply the "factor" function to all columns of the database using the "lapply" function. Lastly, "str(df)" is used again to confirm the correct reformatting.

### 1.5    Remove Unnecessary Columns
Your imported data may contain extra columns with variables that are irrelevant to the current regression problem, such as patient numbers, names, or other endpoints. The latter is the case in the Glioblastoma database: The 22nd column contains the variable "TwelveMonths", which is a binary form of our continuous endpoint of interest "Survival". Leaving this redundant variable in would lead to data leakage – the model could simply take the binary "TwelveMonths" variable and extrapolate some parts of our endpoint "Survival" from it, without actually learning any features. Columns can be removed from a R dataframe by specifying the column number to be removed. "TwelveMonths" is situated in the $22^{nd}$ column of the database, and can thus be removed by applying the function "df <- df[,-22]".

*1.6      Enable Multicore Processing*

If you are working on a machine with multiple central processing unit (CPU) cores, you can enable parallel computing for some functions in R. Using the code in section 1.6, create a computational cluster by entering the number of cores you want to invest into model development.[12] The default is set to 4, a common number of CPU cores in 2020.

*1.7      Partition the Data for Training and Testing*

Figure 1 illustrates the procedure. To randomly split the data into 80% for training an 20% for testing (internal validation), we first set a random seed, such as "set.seed(123)", although you could choose any number. Setting seeds initializes random functions in a constant way, and thus enables reproducibility. Subsequently, we randomly sample 80% of patients and allocate them to the training set ("train"), and do the same for the test set ("test"). Then, the rows of the two newly partitioned sets are shuffled, and the two sets are checked for an approximately equal distribution of the continuous endpoint using "hist(train$Survival)". The histograms show a very similar distribution, both with mean survival of around 12 months.

*1.8      Impute Missing Data*

The glioblastoma database contains no missing data, as the function "VIM::kNN"[23] will let you know. However, should you encounter missing data, this code block should automatically impute missing data using a *k*-nearest neighbor (KNN) algorithm.[1] It is important only to impute missing data within the training set, and to leave the test set alone. This is to prevent data leakage. Also, imputation can be achieved using many different algorithms. We elected to use a KNN imputer for reasons of consistency - during model training, a separate KNN imputer will be co-trained with the prediction model to impute any future missing data.

*1.9      Variable Selection using Recursive Feature Elimination*

Recursive Feature Elimination (RFE) is just one of various methods for variable selection (c.f. Part II for further explanation). In this example, we apply RFE (Figure 2) due to its relative simplicity, generalizability, and reproducibility. Because random functions are involved, seeds need to be set. A linear model is selected as the regressor, and bootstrap resampling with 25 repetitions is used to ensure generalizability of the results. Using the "sizes" argument in the "rfe" function [11], the number of combined variables that are to be assessed can be limited. As we have 20 independent variables, we choose to limit the search for the optimal number and combination of variables to between 10 and 20. The "rfe()" function is executed, which may take some minutes. Using "plot()", the results of the RFE procedure can be illustrated (Figure 3), and it is clear that a combination of 16 variables led to the highest performance. The selected variables are stored in "predictors(RFE)". These 16 selected variables, plus the endpoint "Survival" are stored in "keepvars", and the training set is subsequently reduced to 17 columns.

*1.10     Get a Final Overview of the Data*

Before diving directly into model training, it is advisable to look over the training and test set using the "summary()" function to assess the correctness of the independent variables and the endpoint.

## 2     Model Training

*2.5      Setting up the Training Structure*

Now that the data are prepared, training of the different models can be initiated. In this example, we elected to train five different algorithms to predict continuous survival in months: Linear regression using a generalized linear model (GLM), random forests [2] (RF), generalized additive models [8] (GAM), Least Absolute Shrinkage and Selection Operator (Lasso) regression [9], and ridge regression [9]. A brief overview of the five different models is provided in Table 2. We specifically refrained from using more complex models, such as neural network regressors, due to their inherently decreased interpretability and because they are more prone to overfitting on the relatively simple, clinical data used in this example.[7] All five models are trained sequentially and in a similar way using a universal wrapper that executes training, the "caret" package.[11] Hyperparameters – if available – are tuned automatically. To prevent overfitting, 5-fold cross validation was chosen as resampling technique in this example (Figure 4).[20] However, bootstrap resampling with 25 repetitions could also easily be implemented (c.f. Part IV).

## 2.6     Model Training

The procedure (Figure 4) is equivalent for all five regressors (Sections 2.2.1 to 2.2.5). First, a seed is set to initialize the random number generator in a reproducible way. Subsequently, the algorithm to be used is specified in the "method" argument – the first model to be trained is a linear GLM, so "method = "glm"" is chosen. The "tuneLength" argument depends on the complexity and of the hyperparameters: GLM has no hyperparameters, so a low value is specified. We specify that the parameters and hyperparameters are to be optimized according to root mean square error (RMSE, metric = "RMSE"), and that a KNN imputer is co-trained for future missing data (preProcess = "knnImpute"). The inputs are automatically centered and scaled by the "caret" package. After running the fully specified "caret::train" function, it may take some minutes for all resamples to finish training. The red "STOP" dot at the top right of the RStudio console will be present for as long as the model is training. Subsequently, the resampled performance metrics RMSE, mean average error (MAE), and $R^2$ are calculated. Finally, the model specifications and resampled training performance are printed, and the model can be saved using the "save()" function for potential further use.

After completion of training the GLM (Section 2.2.1), the same procedure is repeated for the GAM (Section 2.2.2), Lasso regressor (Section 2.2.3), ridge regressor (Section 2.2.4), and RF (Section 2.2.5) models.

## 3     Model Evaluation and Selection

### 3.5     Model Training Evaluation

As soon as all five models have been trained, their performance on the training data can be compared. The final model should be selected based upon training data only. Criteria for clinical prediction model selection may include discrimination and calibration on the training set, as well as the degree of interpretability of the algorithm. Section 3.1 compiles the results of all five models, and allows their comparison in terms of RMSE, MAE, and $R^2$. The code in this section will also open a new plot viewer window using "dev.new()", that allows graphical comparison of the five models. If you have executed all parts of the script correctly up to this point, you will be presented with a plot that is identical to Figure 5. In this plot, we see that – while all models performed admirably – the GLM (linear model), GAM, and ridge regressor had the lowest error values (RMSE and MAE). Models perform well if these error values approach 0. In addition, all models except for the RF had very high $R^2$ values, indicating high correlation of predicted with actual survival values. The $R^2$ value, taken together with quantile-quantile plots that will be demonstrated further on, can serve a role similar to calibration measures in binary classification models – namely, as an indication of how well the predicted values correspond to the actual values over the spectrum of survival lengths.[21] As all of the best-performing algorithms are highly interpretable, the GLM, GAM, and ridge regressor would all make fine options for a final model. In this example, we elected to carry on with the ridge regressor.

### 3.6     Select the Final Model

The fully trained ridge regressor was previously stored as "ridgefit", and its resampled training evaluation as "RIDGE" in section 2.2.4. Now, as the ridge regressor model is selected as the final model, it is renamed "finalmodel", and its training evaluation is renamed "finalmodelstats". You can choose any other model by replacing these two terms with the corresponding objects from section 2.2.

### 3.3 Internal Validation on the Test Set

For the first time since partitioning the original Glioblastoma database, the 20% of patients allocated to the test set are now used to internally validate the final model. First, a prediction is made on the test set using "finalmodel" and the "predict()" function. The predicted survival values for the entire test set are then contrasted with the actual survival values from the endpoint (test\$Survival) to arrive at error values. Using "print(Final)", the internal validation metrics can be viewed. Performance that is on par with or slightly worse than the training performance usually indicates a robust, generalizable model. Performance that is relevantly worse than the training performance indicates overfitting during training. These problems are discussed in detail in Part II. The final model can be saved, and will be available as "FINALMODEL.Rdata" in the same folder as the R script. Using the "load()" function, models can be imported back into R at a later date.

If you end up with the same performance metrics for the final ridge regressor as in Table 3, you have executed all steps correctly.

## 4 Reporting and Visualization

When generating clinical prediction models and publishing their results, there is a minimum set of information that ought to be provided to the reader. First, the training methods and exact algorithm type should be reported, if possible along with the code that was used for training. Second, the characteristics of the cohort that was used for training should be provided, such as in Table 1. If multiple cohorts are combined or used for external validation, the patient characteristics should be reported in separate. For regression models, a minimum of RMSE, MAE, and $R^2$ should be reported for both training and testing performance. There are countless other metrics to describe regression performance of clinical prediction models. Lastly, whenever feasible, an attempt at interpreting the model should be made. For example, logistic regression (GLM) models produce odds ratios, and GAMs can produce partial dependence values. However, there are also universal methods to generate variable importance measures that can apply to most regression models, which we present below. To simplify reporting, this final section helps compile all these data required for publication of clinical prediction models. For further information on reporting standards, consult the transparent reporting of a multivariable prediction model for individual prognosis or diagnosis (TRIPOD) checklist.[5]

### 4.5 Compiling Training Performance

The resampled training performance can be printed using "print(finalmodelstats)". The metrics that are produced include RMSE, MAE, and $R^2$. Subsequently, a quantile-quantile (Q-Q) plot is generated for the training set using the "qqplot()" function. A quantile-quantile plot plots quantiles of predicted values against quantiles of true survival values, and can thus be used to judge how a regressor performs over the wide span of survival values (short-term and long-term survivors).

### 4.6 Compiling Internal Validation Performance

Similarly, the performance on the test set (internal validation) can be recapitulated, and a quantile-quantile plot produced (analogous to Figure 6).

### 4.7 Assessing Variable Importance

By using "varImp(finalmodel)", a universal method for estimation of variable importance based on AUC is executed, and results in a list of values ranging from 0 to 100, with 100 indicating the variable that contributed most strongly to the predictions, and vice versa. Finally, "plot(imp)" generates a variable importance plot that can also be included in publication of clinical prediction models (See Figure 7).

## Conclusion

This section presents one possible and standardized way of developing clinical prediction models for regression problems such as patient survival. Proper visualization and reporting of machine learning-based clinical prediction models for continuous endpoints are also discussed. We provide the full, structured code, as well as the complete Glioblastoma survival database for the readers to download and execute in parallel to this section. The methods presented can and are in fact intended to be extended by the readers to new datasets, new endpoints, and new algorithms.

## Disclosures

Funding: No funding was received for this research.

Conflict of interest: All authors certify that they have no affiliations with or involvement in any organization or entity with any financial interest (such as honoraria; educational grants; participation in speakers' bureaus; membership, employment, consultancies, stock ownership, or other equity interest; and expert testimony or patent-licensing arrangements), or non-financial interest (such as personal or professional relationships, affiliations, knowledge or beliefs) in the subject matter or materials discussed in this manuscript.

Ethical approval: All procedures performed in studies involving human participants were in accordance with the ethical standards of the 1964 Helsinki declaration and its later amendments or comparable ethical standards.

Informed consent: No human or animal participants were included in this study.

Figures

```
set.seed(123)    #Setting seeds allows reproducibility
dt = sort(sample(nrow(df), nrow(df)*.80)) #Allocate 80% split
train <-df[dt,]
test <-df[-dt,]
train <- train[sample(nrow(train)),] #Shuffle
test <- test[sample(nrow(test)),]
```

Figure 1 Code section 1.7: This section illustrates how to partition a database into training and test (internal validation) sets.

```
set.seed(123) #Set a seed
rfecontrol <- rfeControl(functions = caretFuncs, method = "boot", number = 25, rerank = F,
          verbose = T, allowParallel = T)
set.seed(123)
RFE <- rfe(Survival ~ ., data = train, sizes=c(10:ncol(train)-1), rfeControl=rfecontrol,
          method = "glm", trControl = trainControl(method = "boot"))
print(RFE) #Results
plot(RFE, type=c("g", "o")) #Plot performance over #vars
predictors(RFE) #variables to be kept
```

Figure 2 Code section 1.9: This section illustrates the recursive feature elimination (RFE) procedure. A generalized linear model (GLM) is chosen as the regressor, along with bootstrap resampling with 25 repetitions.

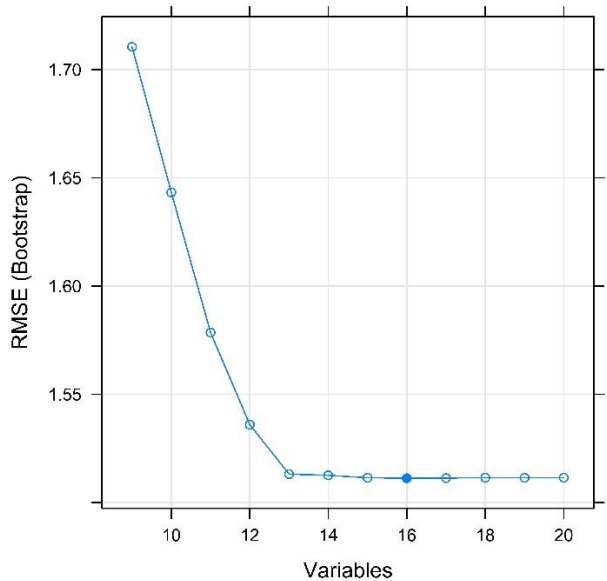

Figure 3 Results of the recursive feature elimination (RFE) variable election procedure. It was determined that using 16 variables explained the highest amount of variance, as seen in the low RMSE that was achieved with this number and combination of variables.

```
#2 MODEL TRAINING ----
#2.1 Set up the Training and Resampling Structure ----
#As a standard, we choose 5-fold cross validation as the resampling method.
ctrl <- trainControl(method = "cv", number = 5, allowParallel = T, returnResamp = "all")

#2.2.1 Train Model #1: (Generalized) Linear Model (LM) ----
set.seed(123)
lmfit <- caret::train(Survival ~ ., data = train, method = "glm", trControl = ctrl, tuneLength = 25, metric = "RMSE", preProcess = "knnImpute", na.action = na.pass)
#we automatically tune our models based on RMSE, and co-train a kNN imputer that then automatically imputes any potential missing data for future samples.
LM <- as.data.frame(cbind(min(lmfit$results$RMSE), min(lmfit$results$MAE), max(lmfit$results$Rsquared)))
colnames(LM)  <- c("RMSE", "MAE", "R2")
#Summary
print(lmfit)
print(LM)
#Save Model File
save(lmfit, file= "LM.Rdata")
```

Figure 4 Code sections 2.1 & 2.2: First, the training structure is established: 5-fold cross validation is used. Subsequently, a linear regression model (generalized linear model, GLM) is trained. All predictor variables are provided to the model, and it is automatically tuned for root mean square error (RMSE). A k-nearest neighbor imputer is co-trained to impute any potential missing data in future predictions. Subsequently, performance is assessed, and the final model information and resampled training performance are printed.

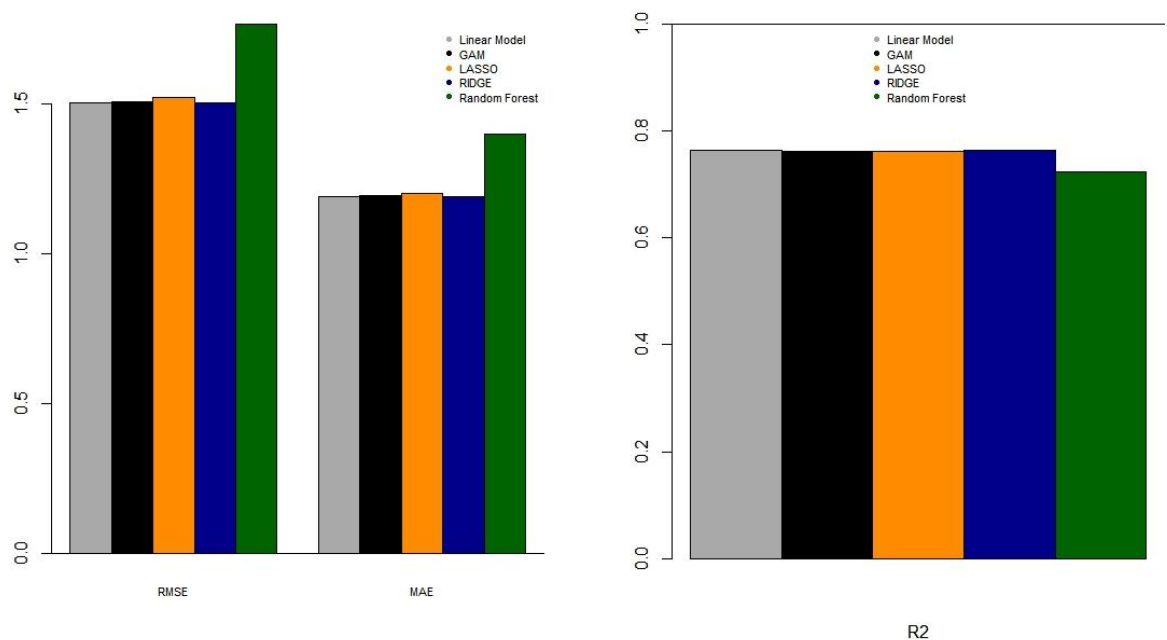

Figure 5 Graphical comparison of root mean square error (RMSE) and mean average error (MAE) to the left, and $R^2$ to the right (Code section 3.1). The linear model, the LASSO model, and the ridge regressor all exhibited similarly low error values (RMSE and MAE), and all three achieved high $R^2$ values.

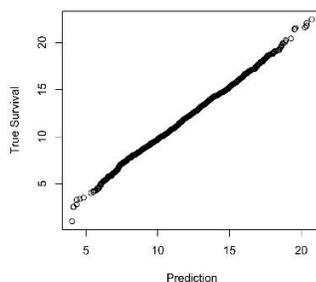

Figure 6 Quantile-Quantile plot for the final ridge regressor, demonstrating the relationship between predicted survival values and actual survival in months on the test set (internal validation). The curve can be interpreted similarly to a calibration curve seen for binary classification models. The curve closely approximates a diagonal line, indicating excellent performance for both short-term and long-term survivors.

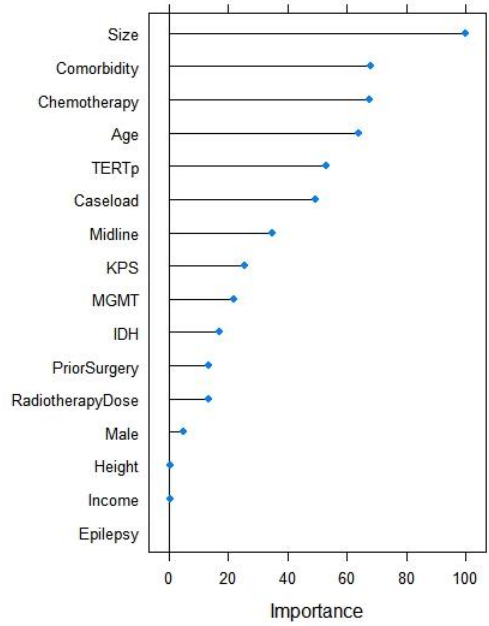

Figure 7 Variable importance of the final model based on a nonparametric, model-independent method. The importance metrics are scaled from 0 to 100.

Tables

Table 1 Structure of the simulated glioblastoma dataset. The number of included patients is 10'000. Values are provided as means and standard deviations or as numbers and percentages.

| Variable Name | Description | Value |
|---|---|---|
| Survival | Overall survival from diagnosis in months | 12.1 ± 3.1 |
| TwelveMonths | Patients who survived 12 months or more from diagnosis | 5184 (51.8%) |
| IDH | IDH mutation present | 4136 (41.4%) |
| MGMT | MGMT promoter methylated | 5622 (56.2%) |
| TERTp | TERTp mutation present | 5108 (51.1%) |
| Male | Male gender | 4866 (48.7%) |
| Midline | Extension of the tumor into the midline | 2601 (26.0%) |
| Comorbidity | Presence of any systemic comorbidity such as diabetes, coronary heart disease, chronic obstructive pulmonary disease, etc. | 5135 (51.4%) |
| Epilepsy | Occurrence of an epileptic seizure | 3311 (33.1%) |
| PriorSurgery | Presence of prior cranial surgery | 5283 (52.8%) |
| Married | Positive marriage status | 5475 (54.8%) |
| ActiveWorker | Patient is actively working, i.e. not retired, student, out of work, etc. | 5459 (54.6%) |
| Chemotherapy | Patients who received chemotherapy for glioblastoma | 4081 (40.8%) |
| HigherEducation | Patients who received some form of higher education | 4209 (42.1%) |
| Caseload | Yearly glioblastoma microsurgery caseload at the treating center | 165.0 ± 38.7 |
| Age | Patient age at diagnosis in years | 66.0 ± 6.2 |
| RadiotherapyDose | Total radiotherapy absorbed dose in Gray | 24.8 ± 6.7 |
| KPS | Karnofsky Performance Scale | 70.5 ± 8.0 |
| Income | Net yearly household income in US dollars | 268'052 ± 62'867 |
| Height | Patient body height in cm | 174.6 ± 6.7 |
| BMI | Deviation of body mass index from 25; in kg/m2 | 0.02 ± 1.0 |
| Size | Maximum tumor diameter in cm | 2.98 ± 0.55 |

Table 2 Overview of the five models that were employed.

| Model | caret::train() input | Package | Suitability | Hyperparameters |
|---|---|---|---|---|
| Generalized Linear Model | glm | stats | Classification, Regression | None |
| Random Forest | rf | randomForest | Classification, Regression | mtry (number of variables at each tree node) |
| Least Absolute Shrinkage and Selection Operator (Lasso) | lasso | elasticnet | Regression | fraction (sum of absolute values of the regression coefficients) |
| Ridge Regression | ridge | elasticnet | Regression | lambda (shrinkage factor) |
| Generalized Additive Model | gamLoess | gam | Classification, Regression | span (smoothing span width), degree (degree of polynomial) |

Table 3 Performance metrics of the final regression model (ridge regression) for glioblastoma survival in months. The difference in performance among training and testing is minimal, demonstrating a lack of overfitting at internal validation.

| Metric | Cohort | |
|---|---|---|
| | Training (n = 8000) | Internal Validation (n = 2000) |
| Root mean square error (RMSE) | 1.504 | 1.515 |
| Mean absolute error (MAE) | 1.191 | 1.211 |
| $R^2$ | 0.763 | 0.759 |

Supplementary Material

Please download the supplementary files from https://micnlab.com/files/

Supplement 1 R Code for binary classification of twelve-month survival of glioblastoma patients.

Supplement 2 Database of 10'000 simulated glioblastoma patients, intended for use with the R code